\documentclass[12pt,onecolumn,twoside]{IEEEtran}
\pdfoutput=1 
\usepackage{graphicx}
\usepackage[english]{babel}
\usepackage{amssymb}
\usepackage{pmat}
\usepackage{balance}
\usepackage{subfigure}
\usepackage{multirow}
\usepackage{longtable}
\usepackage{amsthm}
\usepackage{amssymb }
\usepackage{amsmath}
\usepackage{bm}
\usepackage{mathrsfs}
\usepackage{amsfonts}
\usepackage{longtable}
\usepackage{multirow}
\usepackage{algorithm, algpseudocode}
\usepackage{cite}
\usepackage{stfloats}

\global\long\def\PointOne{P_{1}^{I,J}}
\global\long\def\PointTwo{P_{2}^{I,J}}
\global\long\def\PointThree{P_{3}^{I,J}}
\global\long\def\PointFour{P_{4}^{I,J}}

\global\long\def\ZOne{Z_{1}^{I,J}}
\global\long\def\ZTwo{Z_{2}^{I,J}}
\global\long\def\ZThree{Z_{3}^{I,J}}
\global\long\def\ZFour{Z_{4}^{I,J}}

\newcommand{\myvec}[2]{\vec{#1}_{#2}}
\newcommand{\cellij}{C_{I,J}}

\global\long\def\Cell{I,J}

\global\long\def\FrameCell{\mathcal{F_{\textrm{C}_{\textrm{I,J}}}}}
\newcommand{\DelZ}[1]{\Delta Z_{#1}^{I,J}}

\title{Dynamic Endpoint Object Conveyance Using a Large-Scale Actuator Network}
\author{
\authorblockN{Martin Sinclair, Ioannis A. Raptis}\\
\authorblockA{Department of Mechanical Engineering, University of Massachusetts Lowell\\
  Email: $\{$Martin\_Sinclair, Ioannis\_Raptis$\}$@uml.edu}

}

\begin{document}

\maketitle

\begin{abstract}
Large-Scale Actuator Networks (LSAN) are a rapidly growing class of
electromechanical systems. A prime application of LSANs in the industrial
sector is distributed manipulation. LSAN's are typically implemented
using: vibrating plates, air jets, and mobile multi-robot teams. This
paper investigates a surface capable of morphing its shape using an
array of linear actuators to impose two dimensional translational
movement on a set of objects. The collective nature of the actuator
network overcomes the limitations of the single Degree of Freedom
(DOF) manipulators, and forms a complex topography
to convey multiple objects to a reference location. A derivation of the
kinematic constraints and limitations of an arbitrary multi-cell surface
is provided. These limitations determine the allowable actuator alignments
when configuring the surface. A fusion of simulation and practical
results demonstrate the advantages of using this technology over static
feeders.
\end{abstract}

\section{Introduction}

Large-Scale Actuator Networks (LSANs) are systems that are comprised of interconnected
and spatially distributed simplistic actuators. These actuators work
as a collective to complete global objectives that are far beyond
the capabilities of the individual components. By using these systems,
the need for expensive and complex single purpose mechanisms is becoming
obsolete. The cooperative nature of the network provides a robust and scalable
mechanism that is capable of functioning despite the failure of individual
elements.

The predominant application of LSANs is distributed manipulation.
This field involves the conveyance of objects using a large number
of individual manipulators. Many different types of actuators have
been used in the development of scalable distributed manipulators. The individual
components that create an LSAN are typically simplistic actuators, i.e. air jets or linear/rotary actuators. Recent advances in MEMS (Micro Electro Mechanical Systems) technology led to the inexpensive fabrication of micro manipulators; allowing the placement of large number of actuating elements in smaller areas. This high manipulator density results to improved actuation resolution.

There are many examples of the current state of the art in the field
of distributed manipulation. A typical example is vibrating plates part feeders reported in ~\cite{reznik:1998flat,reznik:2000building,bohringer:1996upper,quaid:2000design,beal:2006infrastructure}. Vibratory feeders utilize inclined vibrations to transport parts along a track. In ~\cite{reznik:1998flat,reznik:2000building} parallel manipulation of multiple parts (translation and orientation) is achieved by a single vibrating plate. Programmable vector techniques for vibratory part feeders and MEMS actuator arrays have been extensively studied in ~\cite{bohringer:1996upper,bohringer:1999programmable,bohringer:2000}. These works present an analytic derivation of an artificial force field method that utilize a massive number of microscopic actuators (`motion pixels') in order to apply frictional forces that transport and rotate planar objects to pre-defined locations. MEMS micro-actuators designed for micro manipulation are further investigated in ~\cite{konishi:2000autonomous,donald:2008planar}. These systems implement a very large number of actuators to generate even small motions. The work in ~\cite{yim:2000two} presents ``Polybot'', a modular system used to move non-planar objects by using cilia out-of-plane manipulators. 

Air-jets have been successfully implemented to direct objects as shown in ~\cite{moon:2006distributed,varsos:2005generation,varsos:2006generation,Varsos:2006min,yim:2000two,laurent:2011new,delettre:20112}. Air-jet conveying is preferred for handling objects that are too fragile for traditional part feeders ~\cite{yim:2000two,delettre:2010new,delettre:20112} or easily contaminated ~\cite{laurent:2011new}. These systems implement static and dynamic potential flow fields ~\cite{Varsos:2006min,varsos:2006generation}. The implementation of sinks to control a single object in a fixed plane was investigated in ~\cite{moon:2006distributed}. 

Conveyor belts is one of the most common methods to transport parts to a target location. The system detailed in ~\cite{akella:1996sensorless} uses a single joint robot to move and orient parts in a speed conveyor belt. Conveyor systems are often limited to a small number of final exit locations. An array of wheels was used to provide the force for transporting an object with distributed manipulation in ~\cite{luntz:2001distributed}. Investigation of the implementation and development of an array of soft and compliant actuators to move fragile objects is shown in ~\cite{tadokoro:2000distributed}. Part feeding on a surface with frictional contacts using protrusions is presented in ~\cite{song:2005two}. Other distributed manipulable surfaces are investigated in ~\cite{yu:2010biologically} with a focus on maintaining a level platform despite changing ground conditions. The surface in ~\cite{yu:2008morpho} changes it shape to drive objects in a single axis. The use of a morphing surface as a means to display 3D objects is shown in ~\cite{leithinger:2010relief}. Self-organized systems of actuator networks are investigated in ~\cite{beal:2006infrastructure} with a focus on controlling networks of up to 10,000 nodes.

This paper presents a morphing surface that autonomously adjusts its shape to transport an arbitrary number of objects to a variable reference location. The reconfiguration of the surface's shape is controlled by a grid of vertical linear actuators that adjust its height at given points in space. The multi-actuator grid results to a mess of interconnected rectangular flat cells that can adjust locally their inclination. This inclination transports the objects that lie on top of the cell by regulating their weight components.    

A detailed analysis of the surface kinematics and object dynamics is provided. This analysis includes an explicit derivation of the constraint equations that apply to the surface and determines the total number of available independent control inputs of the system. In addition, the objects equations of motion reveal how the multi-actuator input controls a multi-task process governed by continuous dynamics. This  analysis was used to design control schemes that are sympathetic to the physical constraints of the system. The main contribution of this work is the design of computationally efficient, low complexity control algorithms that comply with the physical constraints of the mechanism and are scalable to systems with a large number of actuators. To validate the applicability of this approach, a prototype system was developed. The prototype demonstrated how simplistic and computationally attractive control algorithms combined with minimalistic hardware can be used to transport objects to multiple directions of the workspace as might be needed on automated assembly lines or in automated warehouses. 

This paper is organized as follows: Section \ref{sec:system-description} provides a detailed mathematical description of the morphing surface. The kinematic analysis of a single cell is given in \ref{sec:kinematics-and-control}. The same Section outlines a preliminary control law for transporting objects over a single cell. The kinematic description of a multi-cell surface is given in Section \ref{sec:multicell-kinematics}. The control algorithms that govern the transportation of multiple objects over an arbitrary surface are described in Section \ref{sub:Multi-Cell-Control}. A description of a prototype surface that was developed to validate the applicability of the proposed mechanism is given Section \ref{sec:prototype}. Finally, both experimental and simulation results are presented in Section \ref{sec:results}. Concluding remarks are given in Section \ref{sec:Conclusion}.

\section{System Description\label{sec:system-description}}
\subsection{Mathematical Notation\label{sub:Mathematical-Notation}}

The abbreviations $C_{t}$, $S_{t}$ and $T_{t}$ with $t\in\mathbb{R}$
represent the trigonometric functions $\cos(t)$, $\sin(t)$, and
$\tan(t)$, respectively. The superscript $T$ indicates the transpose
of a vector or matrix. The operands $||\cdot||$, $|\cdot|$ denote
the Euclidean norm and the $|\cdot|_{1}$ norm of a vector, respectively.

\subsection{Overview\label{sub:System-description}}

The system being studied is composed of an array of cells that form a three dimensional (3-D) surface. Each cell is defined by the tips of four linear actuators that shape at every time instant a 2D tetrahedron (parallelogram) in space. By constraining each cell to form a flat surface, the complexity of the overall system model is reduced. The actuators are normal to the global inertial plane and are justified in an ordered array (grid). The seam between two adjacent cells was ignored in the model since each cell is treated as a two dimensional (2-D) object. 


Each cell is an element of the grid that is uniquely defined by its row and column. For a grid with $n$ columns and $m$ rows, the surface is denoted as $\mathcal{S\left(\mathtt{n,m}\right)}$. Given these dimensions, the total number of cells that make up the surface is $n\times m$. Let $\mathcal{R}=\left\{ 1,\ldots,m\right\} $ and $\mathcal{C}=\left\{ 1,\ldots,n\right\} $ denote the row and column set, respectively. A cell entry in the surface located at the $I\in\mathcal{R}$ row and $J\in\mathcal{C}$ column of the grid is denoted by $C_{I,J}$.

\begin{figure}
\begin{centering}
\includegraphics[scale=0.7]{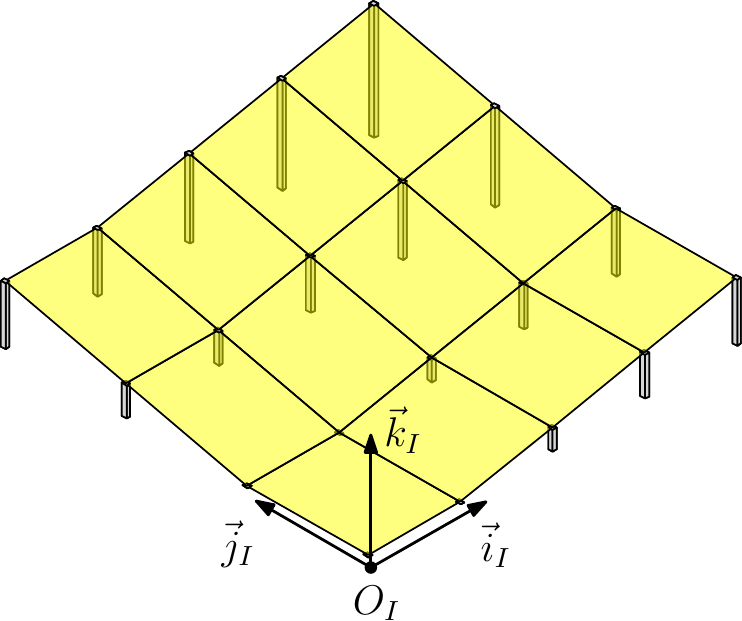}
\par\end{centering}
\caption{The inertial reference frame for an arbitrary surface $S\left(\mathtt{n,m}\right)$.
\label{fig:referance-frames}}
\end{figure}

The coordinates of the actuators were determined with respect to an
inertial fixed frame with its origin located at the lower left actuator
of the cell $C_{1,1}$ when the actuator is retracted to its minimum length. The maximum length of a actuator is denoted by $l$. The inertial frame remains stationary while the actuators change their length with time.
The inertial frame is defined by its origin and three orthonormal
vectors, hence $\mathcal{F}_{I}=\left\{ O_{I},\vec{i}_{I},\vec{j}_{I},\vec{k}_{I}\right\} $. The directions of the vectors $\vec{i}_{I}$ and $\vec{j}_{I}$ can be seen in Figure \ref{fig:referance-frames}, while $\vec{k}_{I}$
points upward such that $\left\{ \vec{i}_{I},\vec{j}_{I},\vec{k}_{I}\right\} $ constitutes a right handed Cartesian coordinate frame $\left(\vec{k}_{I}=\vec{i}_{I}\times\vec{j}_{I}\right)$.
Let $W$ and $L$ denote the distances between two adjacent actuators on each side of a cell measured in the inertial plane $[\vec{i}_{I},\vec{j}_{I}]$. The coordinates of the four actuator tips of cell $C_{I,J}$ are: 
\begin{align}
\PointOne & =\left[\begin{array}{c}
(I-1)\cdot W\\
(J-1)\cdot L\\
\ensuremath{\ZOne}
\end{array}\right] & \PointTwo & =\left[\begin{array}{c}
I\cdot W\\
(J-1)\cdot L\\
\ZTwo
\end{array}\right]\nonumber \\
\PointThree & =\left[\begin{array}{c}
I\cdot W\\
J\cdot L\\
\ZThree
\end{array}\right] & \PointFour & =\left[\begin{array}{c}
(I-1)\cdot W\\
J\cdot L\\
\ensuremath{\ZFour}
\end{array}\right]\label{eq:four-actuators}
\end{align}
In the above equation $Z_{i}^{I,J}$, for $i=1,\ldots,4$, stands for the
coordinate (length) of the $P_{i}^{I,J}$ actuator tip in the $\vec{k}_{I}$
direction of the inertial frame. In order for the object to be transported
over the surface, the actuators must form a flat plane at all time instances. With this constraint, the tips of the actuators will satisfy the following determinant: 
\begin{align}
\left|\begin{array}{c}
\left[\ensuremath{\PointFour}\right]^{T}-\ensuremath{\left[\PointOne\right]}^{T}\\
\left[\ensuremath{\PointFour}\right]^{T}-\ensuremath{\left[\PointTwo\right]}^{T}\\
\ensuremath{\left[\PointFour\right]}^{T}-\left[\ensuremath{\PointThree}\right]^{T}
\end{array}\right|=0\label{eq:3-1}
\end{align}
The above relation holds for any four points that belong to the same flat plane. The enumeration of each cell's actuators starts from the south-west actuator and continuous in a counter clockwise manner. Substituting the coordinates given in \eqref{eq:four-actuators} into \eqref{eq:3-1}, yields the following equation: 
\begin{align}
\ensuremath{\ZThree}=-\ZOne+\ZTwo+\ZFour\label{eq:4-1}
\end{align}
This relation shows that the height of one of the four actuators of each cell is constrained by the height of the other three.  In reality each actuator is regulated using a motor that determines its motion response. Let $Z_{i,com}^{I,J}$ denote the control signals that command the desired length for each of the four linear actuators within a cell. The response of each actuator can be represented satisfactory by a first order differential equation, thus:
\begin{equation}
\tau\dot{Z_{i}}^{I,J}+Z_{i}^{I,J}=Z_{i,com}^{I,J}\;\text{with}\; i=\{1,...,4\},\text{ }I\in\mathcal{C}\text{, }J\in\mathcal{R}
\end{equation}
where $\tau$ is a positive number that represents the time constant of the actuator's motor. A large $\tau$ corresponds to a slow actuator. For the sake of simplicity, in the subsequent analysis
it will be assumed that $Z_{i}^{I,J}$ is the same with $Z_{i,com}^{I,J}$ which allows the actuator dynamics to be disregarded.

\section{Single Cell Kinematics and Control\label{sec:kinematics-and-control}}

\subsection{Cell Kinematics\label{sub:Surface-kinematics}}

\noindent To derive the kinematic equations for the cells, a second
frame of reference is defined. This frame of reference is denoted $\FrameCell=\{O_{\cellij},\myvec{i}{\cellij},\myvec{j}{\cellij},\myvec{k}{\cellij}\}$ with its center $O_{\cellij}$ attached rigidly to the cell $\cellij$. The orthonormal vectors $\myvec{i}{\cellij}$, $\myvec{j}{\cellij}$ change their orientation with time such that they constantly lie on the surface. When all the actuators are leveled, the frames $\mathcal{F}_{I}$ and $\FrameCell$ are aligned to each other. The orientation of cell $C_{I,J}$ at any time may be obtained by performing three consecutive rotations of $\mathcal{F}_{I}$ until the frame is aligned with $\FrameCell$.

\noindent 
\begin{figure}
\begin{centering}
\includegraphics[scale=0.5]{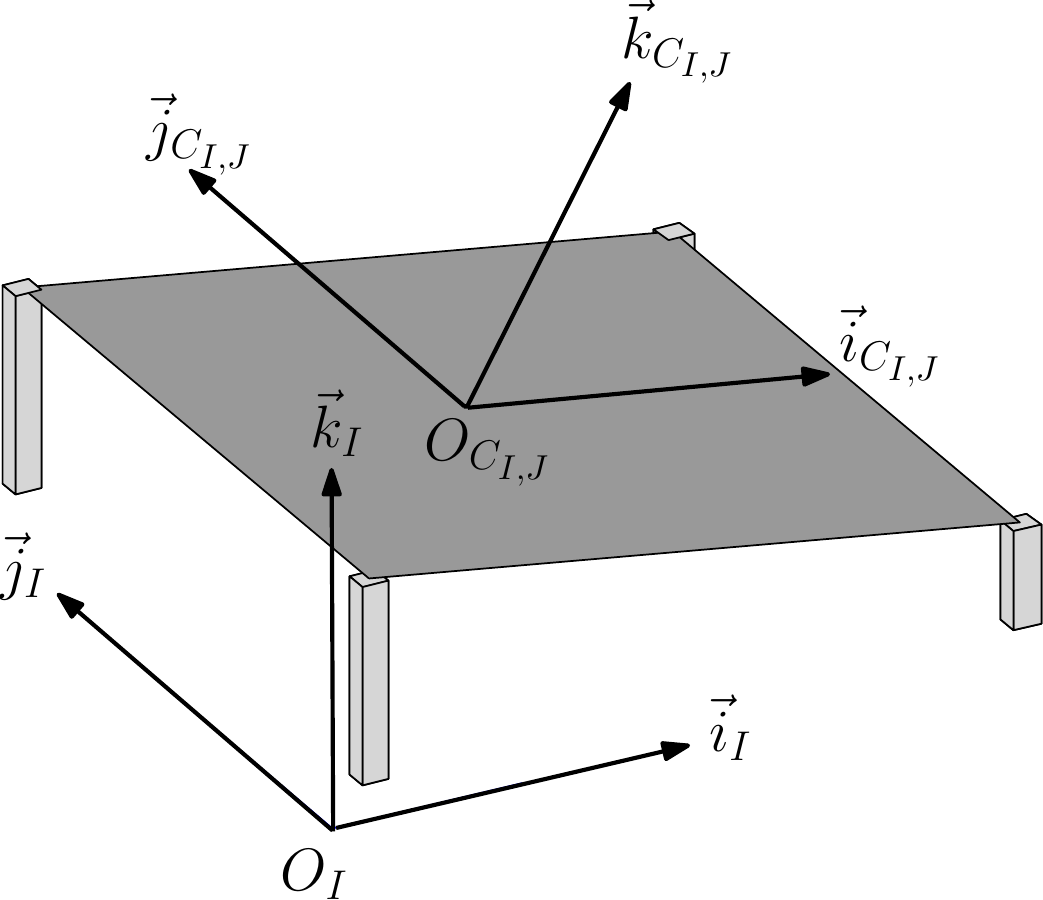} 
\par\end{centering}
\caption{The two reference frames ($\mathcal{F}_{I}$ and $\FrameCell$) for
a single cell configuration.\label{fig:frames}}
\end{figure}

The orientation of $\FrameCell$ is produced by first rotating $\mathcal{F}_{I}$ an angle $\phi_{\Cell}$ about the axis $\vec{i}_{I}$, then rotating by an angle $\theta_{\Cell}$ about $\vec{j}_{I}$, and finally, rotating an angle $\psi_{\Cell}$ about the axis $\vec{k}_{I}$. Using this convention, the angles $\phi_{\Cell}$,$\theta_{\Cell}$, and $\psi_{\Cell}$ are commonly denoted as roll, pitch and yaw angles. Positive direction of each angle is defined by the right-hand rule with respect to the associated axis. Since the cell does not rotate about $\vec{k}_{I}$ we have $\psi_{\Cell}=0$.

The rotation matrix is a systematic way to express the relative
orientation of the two frames. The rotation matrix $R_{\Cell}$ is
parametrized with respect to the three angles: roll ($\phi_{\Cell}$),
pitch ($\theta_{\Cell}$), and yaw ($\psi_{\Cell}$) and it maps vectors
from the cell fixed frame $\FrameCell$ to the inertia frame $\mathcal{F}_{I}$.
Considering that $\psi_{\Cell}=0$ it can be seen that:
\begin{equation}
R_{\Cell}=\left[\begin{array}{ccc}
C_{\theta_{\Cell}} & S_{\theta_{\Cell}}S_{\phi_{\Cell}} & C_{\phi_{\Cell}}S_{\theta_{\Cell}}\\
0 & C_{\phi_{\Cell}} & -S_{\phi_{\Cell}}\\
-S_{\theta_{\Cell}} & C_{\theta_{\Cell}}S_{\phi_{\Cell}} & C_{\theta_{\Cell}}C_{\phi_{\Cell}}
\end{array}\right]\label{eq:rotation_matrix_euler}
\end{equation}
When the rotation matrix is parametrized by the roll-pitch-yaw angles,
singularities occur at $\theta_{\Cell}=\pm\pi/2$. In this system
these singularities are not observed since the actuator limited lengths
prevent the surface from extending to this extreme orientation.

The orientation of each cell is changed by controlling the four corner actuators. By construction, the columns of the orientation matrix express the coordinates of the basis vectors $\{\myvec{i}{\cellij},\myvec{j}{\cellij},\myvec{k}{\cellij}\}$ with respect to the inertial frame. Thus: 
\begin{equation}
R_{\Cell}=\left[\begin{array}{ccc}
i_{C_{I,J}}^{I} & j_{C_{I,J}}^{I} & k_{C_{I,J}}^{I}\end{array}\right]\label{eq:rotation_matrix_vectors}
\end{equation}
In the above equations, the superscript indicates the reference frame
that the basis vector is expressed with respect to. Therefore, in
order to associate $\ZOne,\:\ZTwo$ and $\ZThree$ with $\phi_{\Cell},\theta_{\Cell}$ it is necessary to express the right-hand side of \eqref{eq:rotation_matrix_vectors} with respect to the actuator tip coordinates and then equate the entries of \eqref{eq:rotation_matrix_euler} with \eqref{eq:rotation_matrix_vectors}. The first step is to define
the basis vectors of $\FrameCell$. By definition, the basis vectors $\myvec{i}{\cellij}$, $\myvec{j}{\cellij}$ of the cell-fixed frame lie in the cell's plane. This plane can be uniquely defined by the two vectors $\iota^{I}=\left[\begin{array}{ccc}
W & 0 & \ZTwo-\ZOne\end{array}\right]^{T}$and $\eta^{I}=\left[\begin{array}{ccc}
0 & L & \ZFour-\ZOne\end{array}\right]^{T}$. These vectors are also the two sides of the cell that connect actuators $1$ and $2$ and actuators $1$ and $4$. Since the two vectors are not orthonormal for every $Z_{i}^{\Cell}\in\left[\begin{array}{cc} 0 & l\end{array}\right]$ with $i=1,\ldots,4$, the basis vectors for the cell-fixed frame have to be defined by using the unit vector $i_{C_{I,J}}^{I}=\iota^{I}/||\iota^{I}||$. By definition $\vec{k}_{C_{I,J}}$
is normal to both $\vec{\iota}$ and $\vec{\eta}$, thus by choosing $k_{C_{I,J}}^{I}=\kappa^{I}/||\kappa^{I}||$ with: 
\begin{equation}
\kappa^{I}=\left[\begin{array}{ccc}
\frac{\ZOne-\ZTwo}{W} & \frac{\ZOne-\ZFour}{L} & 1\end{array}\right]^{T}
\end{equation}
the system satisfies the condition $\iota^{I}\cdot\left(\kappa^{I}\right)^{T}=\eta^{I}\cdot\left(\kappa^{I}\right)^{T}=0$.
Equating the (1,1), (3,1) and (2,3), (3,3) components of $R_{I,J}$,
the following kinematic relations can be derived: 
\begin{align}
S_{\theta_{\Cell}} & =\frac{\DelZ{1}}{W}C_{\theta_{\Cell}}\label{eq:kinematic_1}\\
-S_{\phi_{\Cell}} & =\frac{\DelZ{2}}{L}C_{\theta_{\Cell}}C_{\phi_{\Cell}}\label{eq:kinematic_2}
\end{align}
where $\DelZ{1}=\ZOne-\ZTwo$ and $\DelZ{2}=\ZOne-\ZFour$. These equations are used to relate the coordinates of the actuator tips with the orientation angles of the plane to determine the total number of available independent control variables.

\subsection{Object Dynamics\label{sub:Object-Dynamics}}

\begin{figure}
\centering
\includegraphics[width=1.75in]{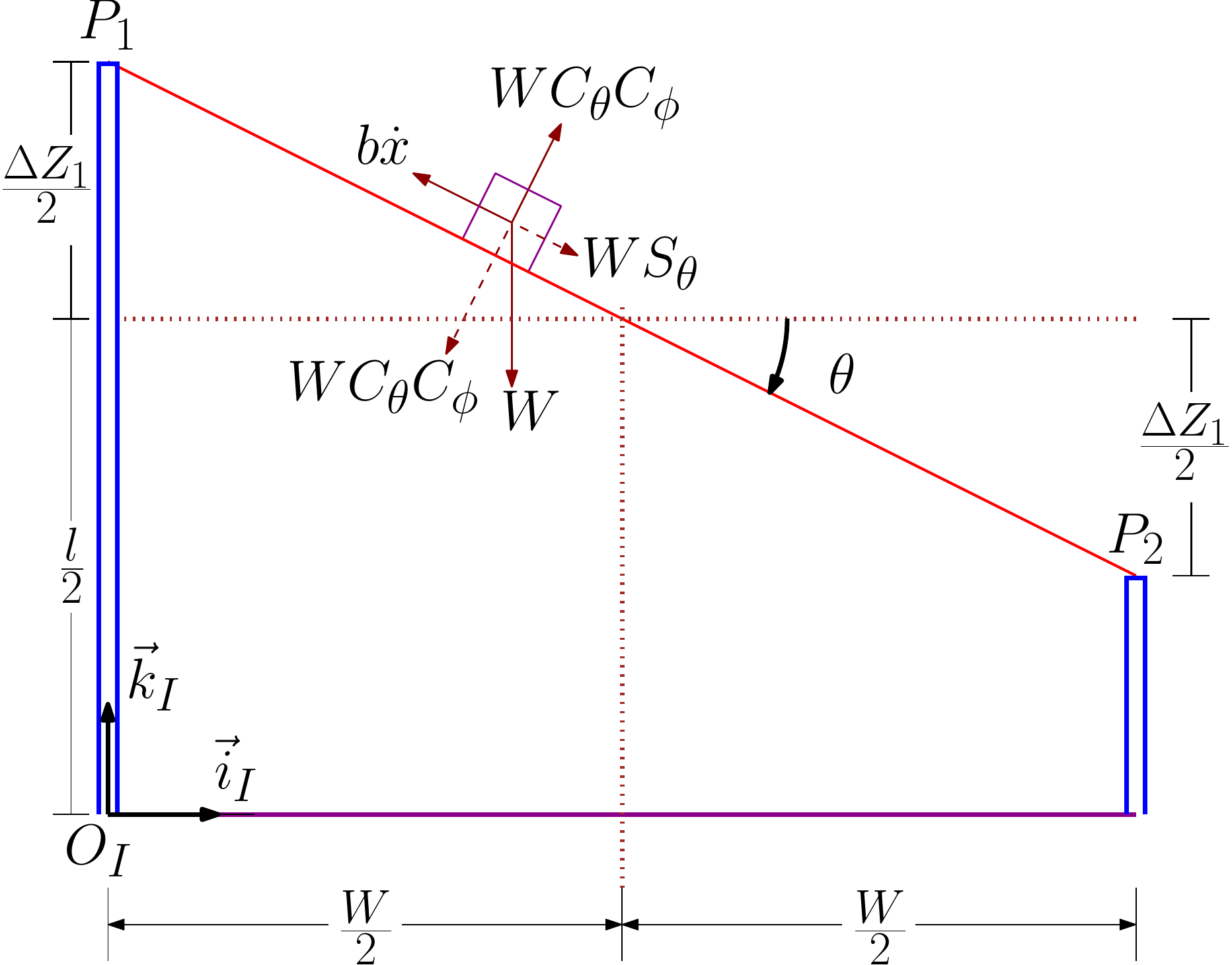}\includegraphics[width=1.75in]{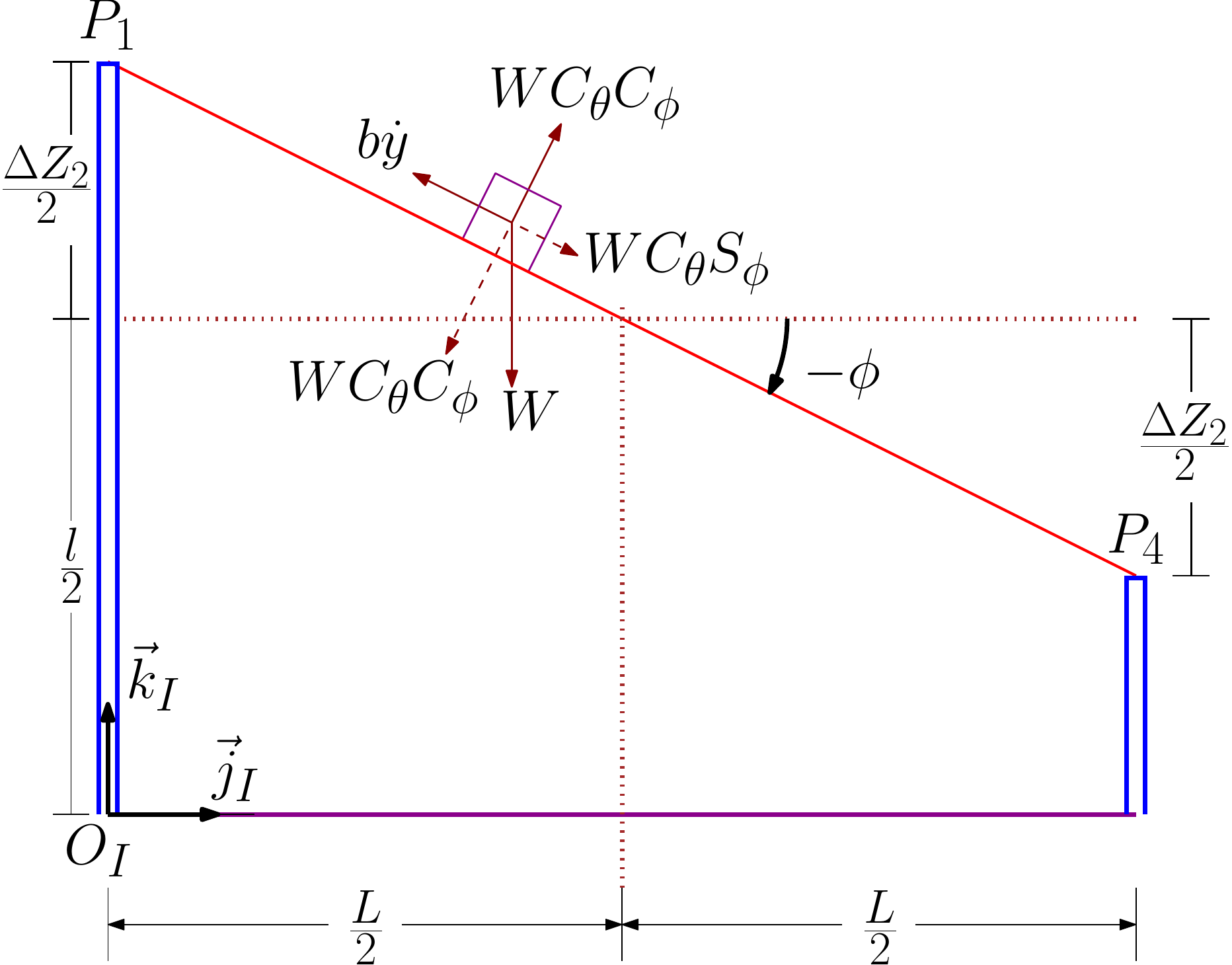}
\caption{The two side views of a single cell with the force components that act on the object.\label{fig:forces}}
\end{figure}

\noindent The object's position with respect to the inertial frame
is denoted by $p^{I}$=$\left[\begin{array}{ccc}
x & y & z\end{array}\right]^{T}\in\mathbb{R}^{3}$. The motion analysis is restricted over a single cell surface ($\mathcal{S}(\mathtt{1},\mathtt{1})$).
As the relations derived in this Section refer to all cells, the indices
that correspond to the cell number are dropped ($I$, $J$). The net
forces applied to the object with respect to the inertial frame are
denoted by $\Sigma F^{I}$. From Newton's second law, the equations
of motion for the object are: 
\begin{equation}
m\ddot{p}^{I}=\Sigma F^{I}\label{eq:newton}
\end{equation}
where $m$ is the mass of the object. The object on top of the surface
is subject to three forces: i) the components of the object's weight
that lie on the surface plane, ii) the friction that opposes the object's
motion, and iii) the reaction force from the surface that is normal
to the plane, facing upward. The object dynamics are manipulated by
changing the orientation of the surface using the actuators. The weight
of the object (expressed in the inertial frame) is $W^{I}=-\left[\begin{array}{ccc}
0 & 0 & mg\end{array}\right]^{T}.$ Using \eqref{eq:rotation_matrix_euler}, the weight components in
the cell-fixed frame are given by: 
\begin{equation}
W^{C}=R^{T}W^{I}=mg\left[\begin{array}{ccc}
S_{\theta} & -C_{\theta}S_{\phi} & -C_{\theta}C_{\phi}\end{array}\right]^{T}\label{eq:weight}
\end{equation}
The reaction from the surface to the object is opposite and equal
to the weight component in the $\vec{k}_{C}$ direction, thus: 
\begin{equation}
N^{C}=mg\left[\begin{array}{ccc}
0 & 0 & C_{\theta}C_{\phi}\end{array}\right]^{T}\label{eq:normal}
\end{equation}
Finally, the friction force from the surface opposes the motion of
the object and is modeled as: 
\begin{equation}
F^{C}=-bv^{C}\label{eq:friction}
\end{equation}
where $b>0$ is the friction coefficient and $v^{C}\in\mathbb{R}^{3}$
is the velocity of the object expressed in the object-fixed frame.
The component of $v^{C}$ in the $\vec{k}_{C}$ direction is zero
since the object always lies on the surface. Using Newtons second
law, it is easy to show that $F^{I}=-bv^{I}$ where $v^{I}=\left[\begin{array}{ccc}
\dot{x} & \dot{y} & \dot{z}\end{array}\right]^{T}\in\mathbb{R}^{3}$ is the object's velocity expressed in the inertial frame. The components
of the three forces applied to the object are illustrated in Figure
\ref{fig:forces}. Substituting \eqref{eq:weight}-\eqref{eq:friction}
to \eqref{eq:newton} one has: 
\begin{align}
\left[\begin{array}{c}
\ddot{x}\\
\ddot{y}\\
\ddot{z}
\end{array}\right] & =g\left[\begin{array}{c}
C_{\theta}C_{\phi}^{2}S_{\theta}\\
-C_{\theta}C_{\phi}S_{\phi}\\
C_{\theta}^{2}C_{\phi}^{2}-1
\end{array}\right]-b\left[\begin{array}{c}
\dot{x}\\
\dot{y}\\
\dot{z}
\end{array}\right]\label{eq:eom}
\end{align}

\subsection{Elementary Reference Point Control\label{sub:Single-Cell-Control}}

This Section describes a control algorithm that autonomously
transports the object to a reference location of a single cell with
inertial coordinates $\left(x_{r}^{I},y_{r}^{I}\right)$. The reference coordinate
in the $\vec{k}_{I}$ directions is of no importance since the object's
$z^{I}$ coordinate is constrained by the surface orientation.  The
feedback law is translated into height differences of the four actuators'
lengths. 

To proceed with the analysis of the control law, we make a coordinate
change to the error variables $e_{x}=x^{I}-x_{r}^{I}$ and $e_{y}=y^{I}-y_{r}^{I}$.
From \eqref{eq:eom} the error dynamics become: 
\begin{align}
\ddot{e}_{x}+b\dot{e}_{x}+C_{\theta}^{2}C_{\phi}^{2}\frac{g}{W}\Delta Z_{1} & =0\label{eq:error_1}\\
\ddot{e}_{y}+b\dot{e}_{y}+C_{\theta}^{2}C_{\phi}^{2}\frac{g}{L}\Delta Z_{2} & =0\label{eq:error_2}
\end{align}
where $\Delta Z_{1}=Z^{1}-Z^{2}$ and $\Delta Z_{2}=Z^{1}-Z^{4}$. The term
$C_{\theta}^{2}C_{\phi}^{2}$ lies in the $\left(\begin{array}{cc}
0 & 1\end{array}\right]$ interval since $\theta,\phi\in\left(\begin{array}{cc}
-\pi/2 & \pi/2\end{array}\right)$. Therefore, the error dynamics are described by two identical second
order nonlinear differential equations with positive parameters. Any
feedback law of the position error can achieve the control objective.
Due to the limitation of the actuator lengths, we chose the following
saturated feedback functions: 
\begin{equation}
\Delta Z_{1}=K_{x}sat_{W}\,e_{x}  \quad\quad  \Delta Z_{2}=K_{y}sat_{L}\,e_{y}
\end{equation}
where $K_{x},$ $K_{y}$, $M_{x}$ and $M_{y}$ are positive constants.
The saturation function $sat_{M}\left(\cdot\right)$ is defined as:
\begin{align*}
sat_{M}(x) & =\begin{cases}
x & \left|x\right|\leq M\\
sign(x)\cdot M & \textrm{else}
\end{cases}
\end{align*}
Elementary Lyapunov stability arguments can show the convergence of
the above control law. The final step is to configure the actuator
lengths based on the values of $\Delta Z_{1}$ and $\Delta Z_{2}$.
The following choice for the actuators tips will guarantee that the
surface will orient about the axis that lie in the midpoints of the
width and length of the cell: 
\begin{align*}
Z_{1} & =\frac{l}{2}+\frac{\Delta Z_{1}}{2}+\frac{\Delta Z_{2}}{2} & Z_{2} & =\frac{l}{2}-\frac{\Delta Z_{1}}{2}+\frac{\Delta Z_{2}}{2}\\
Z_{3} & =\frac{l}{2}-\frac{\Delta Z_{1}}{2}-\frac{\Delta Z_{2}}{2} & Z_{4} & =\frac{l}{2}-\frac{\Delta Z_{1}}{2}+\frac{\Delta Z_{4}}{2}
\end{align*}
The motion of the object is damped by the friction coefficient $b$.
However, even the in the absence of friction the above controller
can guarantee that $\left(x^{I},y^{I}\right)\rightarrow\left(x_{r}^{I},y_{r}^{I}\right)$.
To inject additional damping to the object's motion, a velocity feedback
term may be added to the control inputs. To re-ensure that the actuators'
tips lie in $\left[\begin{array}{cc}
0 & l\end{array}\right]$ interval the control inputs are bounded by $\left|\Delta Z_{1}\right|,\left|\Delta Z_{2}\right|\leq l/2$.
Therefore, the feedback gains are limited to $K_{x}\leq l/2W$ and
$K_{y}\leq l/2L$.

\section{Multi-Cell Kinematics\label{sec:multicell-kinematics}}

\subsection{Multi-Cell Constraints\label{sub:Inter-Cell-Constraints}}

Each individual cell of a $\mathcal{S}(\mathtt{n},\mathtt{m})$ morphing surface has two Degrees of Freedom (DOF) and may transport the object by changing its pitch and roll. Since adjacent cells share a common pair of actuators, at any given point in time, the length of the actuator tips for each cell has to additionally satisfy \eqref{eq:4-1} so that the surface has no discontinuities. This requirement reduces the total DOF for the system. 

To provide a systematic derivation of the total constraints in an arbitrary $\mathcal{S}(\mathtt{n},\mathtt{m})$ surface, we initially examine the two most simple cases involving a pair of cells arranged in either a vertical or horizontal configuration. 

\subsubsection{Vertical Configuration}
\begin{figure} 
\centering 
\subfigure[$\mathcal{S}\mathtt{\left(1,2\right)}$]{\includegraphics[scale=0.8]{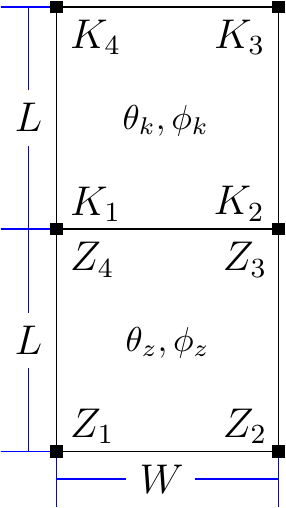}} \subfigure[$\mathcal{S}\mathtt{\left(2,1\right)}$]{\includegraphics[scale=0.8]{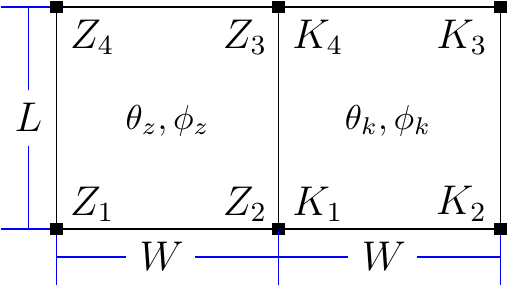}}  
\caption{Two-cell configurations (vertical and horizontal).\label{fig:two-cells}} 
\end{figure}

In the $\mathcal{S}(\mathtt{1},\mathtt{2})$ configuration shown in Figure \ref{fig:two-cells}-(a), the two adjacent cells are constrained by the shared pair of actuators on their common side. Therefore, adjacent cells have the same height profile
along their shared side. For simplicity the two cell are denoted by $C_{Z}$ and $C_{K}$, respectively. 

It can be seen from Figure \ref{fig:two-cells}-(a) that $Z_{4}=K_{1}$ and $Z_{3}=K_{2}$ since they correspond to the same points in space. Since \eqref{eq:4-1} holds true for both cells:
\begin{equation}
K_{2}-K_{1}=Z_{3}-Z_{4}=Z_{2}-Z_{1}\label{eq:Constrant_virt-1}
\end{equation}
Rearranging the terms of \eqref{eq:kinematic_1}, the pitch angle $\theta$
of each cell can be expressed in terms of the actuator heights of
the common side. More specifically:
\begin{align}
T_{\theta_{z}}=\frac{Z_{1}-Z_{2}}{W} &  & T_{\theta_{k}}=\frac{K_{1}-K_{2}}{W}\label{eq:geometry of surface virt}
\end{align}
Substituting \eqref{eq:Constrant_virt-1} into \eqref{eq:geometry of surface virt} provides a direct relationship between the pitch orientation of the
two adjacent cells:
\begin{equation}
T_{\theta_{Z}}=T_{\theta_{K}}\rightarrow\theta_{Z}=\theta_{K}\label{eq:Constrant_orient_virt}
\end{equation}
This constraint means that two cells that are aligned vertically must
share the same pitch angle $\theta$. Since there are no limitations
in the roll orientation of each cell, the $\mathcal{S}(\mathtt{1},\mathtt{2})$
surface has three DOF.

\subsubsection{Horizontal Configuration}
The horizontal configuration of $\mathcal{S}\mathtt{(2,1)}$ shown in Figure \ref{fig:two-cells}-(b),
shows that $Z_{2}=K_{1}$ and $Z_{3}=K_{4}$. Since both cells are required
to form a continuous surface, application of \eqref{eq:4-1} yields: 
\begin{equation}
Z_{2}-Z_{3}=Z_{1}-Z_{4}=K_{1}-K_{4}\label{eq:Constrant_side-1}
\end{equation}
Combining \eqref{eq:kinematic_2} with \eqref{eq:Constrant_side-1},
the relationship between the pitch and roll angles and the common-side actuators can be derived as: 
\begin{align}
\frac{T_{\phi_{Z}}}{C_{\theta_{Z}}}=\frac{Z_{1}-Z_{4}}{L} &  & \frac{T_{\phi_{K}}}{C_{\theta_{K}}}=\frac{K_{1}-K_{4}}{L}\label{eq:First side constrant}
\end{align}
From \eqref{eq:Constrant_side-1} and \eqref{eq:First side constrant}
a nonholonomic constraint is derived that relates the orientation between adjacent cells:
\begin{equation}
\frac{T_{\phi_{K}}}{C_{\theta_{K}}}=\frac{T_{\phi_{Z}}}{C_{\theta_{Z}}}\label{eq:Side constraint orientation}
\end{equation}
The $\mathcal{S}\mathtt{(2,1)}$ surface has has four independent
generalized coordinates describing the orientation of the system ($\theta_{Z}$,$\phi_{Z}$, $\theta_{K}$ and $\phi_{Z}$) and one constraint given
in \eqref{eq:Side constraint orientation}. Similar to the previous
case, this results in a system with three DOF. 

\subsubsection{Square Configuration}
\begin{figure}
\begin{centering}
\includegraphics{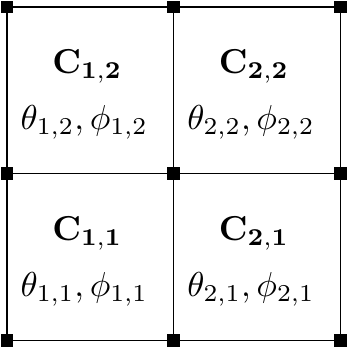}
\par\end{centering}
\caption{A $\mathcal{S}(\mathtt{2},\mathtt{2})$ surface in a square configuration\label{fig:2x2_cells}}
\end{figure}

The two constraint in both directions of the grid will be combined to an illustrative example for the case of a square $\mathcal{S}(2,2)$ surface as shown in Figure \ref{fig:2x2_cells}. From the vertical
alignment:

\begin{equation}
\begin{array}{ccc}
T_{\theta_{1,1}}=T_{\theta_{1,2}} & \longrightarrow & \theta_{1,2}=\theta_{1,1}\\
T_{\theta_{2,1}}=T_{\theta_{2,2}} & \longrightarrow & \theta_{2,2}=\theta_{2,1}
\end{array}\label{eq:Virt_2x2_const-1}
\end{equation}
The constraints equations due to the horizontal placement
are:

\begin{equation}
\begin{array}{ccc}
\frac{T_{\phi_{1,1}}}{C_{\theta_{1,1}}}=\frac{T_{\phi_{2,1}}}{C_{\theta_{2,1}}} & \longrightarrow & T_{\phi_{2,1}}=\frac{C_{\theta_{1,1}}}{C_{\theta_{2,1}}}T_{\phi_{1,1}}\\
\frac{T_{\phi_{1,2}}}{C_{\theta_{1,2}}}=\frac{T_{\phi_{2,2}}}{C_{\theta_{2,2}}} & \longrightarrow & T_{\phi_{2,2}}=\frac{C_{\theta_{1,1}}}{C_{\theta_{2,1}}}T_{\phi_{1,2}}
\end{array}\label{eq:side_2x2_const-1}
\end{equation}

In total, there are eight generalized coordinates determining the
orientation of the $\mathcal{S}(\mathtt{2},\mathtt{2})$ surface.
Due to the constraints of the neighboring cells, the total number
of DOF drops to four. It becomes apparent that the control resources
available to manipulate the surface are a common pitch angle for every
column, and a common roll angle for every row. Every independent
pitch and roll inclination translates to a difference $\Delta\ZOne=\ZOne-\ZTwo$
and $\Delta Z_{2}^{I,J}=\ZOne-Z_{4}^{I,J}$ of the actuators' heights,
respectively. 

\subsubsection{Arbitrary Configuration\label{sub:Arbitrary-Configuration}}
The available DOF for a generic $\mathcal{S}\mathtt{(n,m)}$ surface is determined by the total number of control inputs available to the system. From the vertical configuration described above:
\begin{equation}
T_{\theta_{I,J}}=T_{\theta_{I,1}}\forall\text{ }I\in\mathcal{C}\text{, }J\in\mathcal{R}\rightarrow\theta_{I,J}=\theta_{I,1}=\Theta_{I}\label{eq:pitch constrant-1}
\end{equation}
\noindent The nonholonomic constraint between roll angles in the horizontal configuration has a dependency on the pitch angles which gives the relationship:
\begin{equation}
\begin{array}{ccc}
\frac{T_{\phi_{I,J}}}{C_{\theta_{I,J}}}=\frac{T_{\phi_{I+1,J}}}{C_{\theta_{I+1,J}}} & \longrightarrow & T_{\phi_{I+1,J}}=\frac{C_{\Theta_{I+1}}}{C_{\Theta_{I}}}T_{\phi_{I,J}}\end{array}\label{eq:roll constrant _1}
\end{equation}
\noindent Through iterative calculations it can be shown that for every
row of the grid:
\begin{equation}
T_{\phi_{I,J}}=\frac{C_{\Theta_{I}}}{C_{\Theta_{1}}}T_{\Phi_{J}}\text{where }\Phi_{J}=\phi_{1,J}\text{ and }I\in\left\{ 2,\ldots,n\right\} ,J\in\mathcal{R}
\end{equation}
From (\ref{eq:pitch constrant-1}) and (\ref{eq:roll constrant _1}),
it can be seen that the pitch and roll angles of any cell $C_{I,J}$
in an arbitrary surface $\mathcal{S}\mathtt{(n,\mathtt{m})}$ can
be written as a function of a single row of $n$ pitch angles
$\Theta_{I}$ (with $I\in\mathcal{C}$) , and a single column of $m$ roll
angles $\Phi_{J}$ (with $J\in\mathcal{R}$) . More precisely, for
every cell $C_{I,J}$:
\begin{align}
\theta_{I,J} & =\Theta_{I}\\
\phi_{I,J} & =\arctan\left(\frac{C_{\Theta_{I}}}{C_{\Theta_{1}}}T_{\Phi_{J}}\right)
\end{align}
These relationships show that the total DOF for the surface are $n+m$. More specifically, there are $2\cdot n\cdot m$ generalized coordinates and $2\cdot n\cdot m-n-m$ constraints for the rotational motion of the surface. The DOF dictate the number of independent coordinates that are needed to completely define the orientation of the surface. The inclination of each cell $C_{I,J}$ is determined by the actuators' height differences $\Delta\ZOne=\ZOne-\ZTwo$ and $\Delta Z_{2}^{I,J}=\ZOne-Z_{4}^{I,J}$. The constraints of (\ref{eq:pitch constrant-1}) and (\ref{eq:roll constrant _1}) can be expressed with respect to the height differences $\Delta\ZOne$ and $\Delta Z_{2}^{I,J}$ of each cell. Hence, for every $I\in\mathcal{C}$ and $J\in\mathcal{R}$:
\begin{align}
\Delta\ZOne & =\Delta Z_{1}^{I,1}=\Delta Z_{1}^{I}\\
\Delta Z_{2}^{I,J} & =\Delta Z_{2}^{1,J}=\Delta Z_{2}^{J}
\end{align}
where $\Delta Z_{1}^{I}=Z_{1}^{I,1}-Z_{2}^{I,1}$ and $\Delta Z_{2}^{J}=Z_{1}^{1,J}-Z_{4}^{1,J}$.

\begin{figure}
\begin{centering}
\includegraphics[scale=0.9]{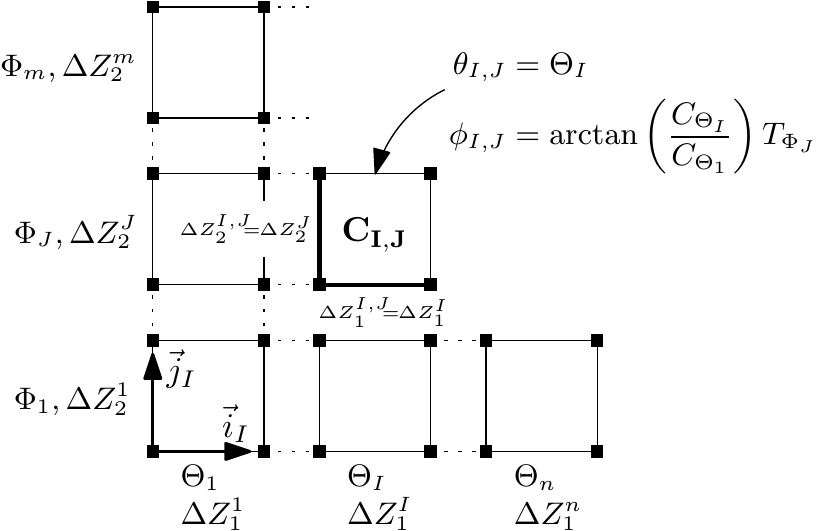}
\par\end{centering}

\caption{This Figure illustrates the independent control variables for each
cell and the constraints of the orientation angles. \label{fig:3x2_cells}}
\end{figure}

The control algorithm is designed to coordinate the values of the $n+m$ control inputs $\Delta Z_{1}^{I}$ ($I\in\mathcal{C}$) and $\Delta Z_{2}^{J}$ ($J\in\mathcal{R}$) such that all of the objects on the surface are transported to the reference cell $C_{I_{r},J_{r}}$. The spatial distribution of the available independent control variables and the constraints in the orientation of each cell can be viewed in Figure \ref{fig:3x2_cells}.

\section{Multi-Cell Control\label{sub:Multi-Cell-Control}}
In Section \ref{sec:multicell-kinematics} it was shown that a surface $\mathcal{S}\mathtt{\left(n,m\right)}$ has $n+m$ DOF, each of which must be specified to uniquely define the overall surface orientation. The objective of the control algorithm is to allocate these DOF in a manner which allows the system to achieve the global goal of moving all objects to the reference cell $C_{I_{r},J_{r}}$ most efficiently.

The control law accomplishes its task by changing $\Delta Z_{1}^{I}=Z_{1}^{I,1}-Z_{2}^{I,1}\left(I\in\mathcal{C}\right)$ and $\Delta Z_{2}^{J}=Z_{1}^{1,J}-Z_{4}^{1,J}$$\left(J\in\mathcal{R}\right)$ defined in Section \ref{sub:Surface-kinematics}. Initially, the height of the target cell is leveled to attain the lowest potential energy. The heights of the rest of the cells are set to be proportional to the distribution of objects between the target cell and the external boundaries of the surface. Two variations of this general logic have been explored and are described in detail. The final step of the collective control law is to translate the inclinations of the cells to height adjustments of the individual actuators of the grid.  

\subsection{Distributed Allocation Control Logic\label{sub:Distributed-Allocation-Control}}
The rational behind this control algorithm is to maximize the utilization of the actuators' limited heights by distributing the inclination of the surface over only the rows and columns that contain objects. It is hypothesized that a dynamic height adjustment will result to faster convergence time than using a static feed with fixed inclination. 

The first step of this algorithm determines the column and row
sets $\bar{\mathcal{C}}\subset\mathcal{C}$ and $\bar{\mathcal{R}}\subset\mathcal{R}$, that correspond to cells that contain objects. Each of these two sets is further broken down into two subsets:
\begin{alignat}{2}
\bar{\mathcal{C}_{l}} & = & \left\{ I\in\bar{\mathcal{C}}|I<I_{r}\right\} \qquad\bar{\mathcal{R}_{d}} & =\left\{ J\in\bar{\mathcal{R}}|J<J_{r}\right\} \nonumber \\
\bar{\mathcal{C}_{r}} & = & \left\{ I\in\bar{\mathcal{C}}|I>I_{r}\right\} \qquad\bar{\mathcal{R}_{u}} & =\left\{ J\in\bar{\mathcal{R}}|J>J_{r}\right\} 
\end{alignat}

If $\bar{C}_{l}$, $\bar{C}_{r}$ and $\bar{R}_{d}$, $\bar{R}_{u}$ denote the cardinality of $\bar{\mathcal{C}}_{l}$, $\bar{\mathcal{C}_{r}}$ and $\bar{\mathcal{R}_{d}}$, $\mathcal{\bar{R}}_{u}$, respectively, then the desired actuator height changes $\Delta Z_{1}^{I}$ ($I\in\mathcal{C}$) and $\Delta Z_{2}^{J}$ ($J\in\mathcal{R}$) are given by:
\begin{equation*}
\Delta Z_{1}^{I}=\begin{cases}
a\frac{l}{\bar{C}_{l}} & \text{if }I\in\bar{\mathcal{C}_{l}}\\
-a\frac{l}{\bar{C}_{r}} & \text{if }I\in\bar{\mathcal{C}_{r}}\\
0 & \text{else}
\end{cases} \quad \Delta Z_{2}^{J}=\begin{cases}
b\frac{l}{\bar{R}_{l}} & \text{if }I\in\bar{\mathcal{R}_{d}}\\
-b\frac{l}{\bar{R}_{r}} & \text{if }I\in\bar{\mathcal{R}_{r}}\\
0 & \text{else}
\end{cases}
\end{equation*}

\noindent where $a$ and $b$ are positive percentiles such that $a+b=1$.

\begin{figure}
\begin{centering}
\includegraphics[scale=0.3]{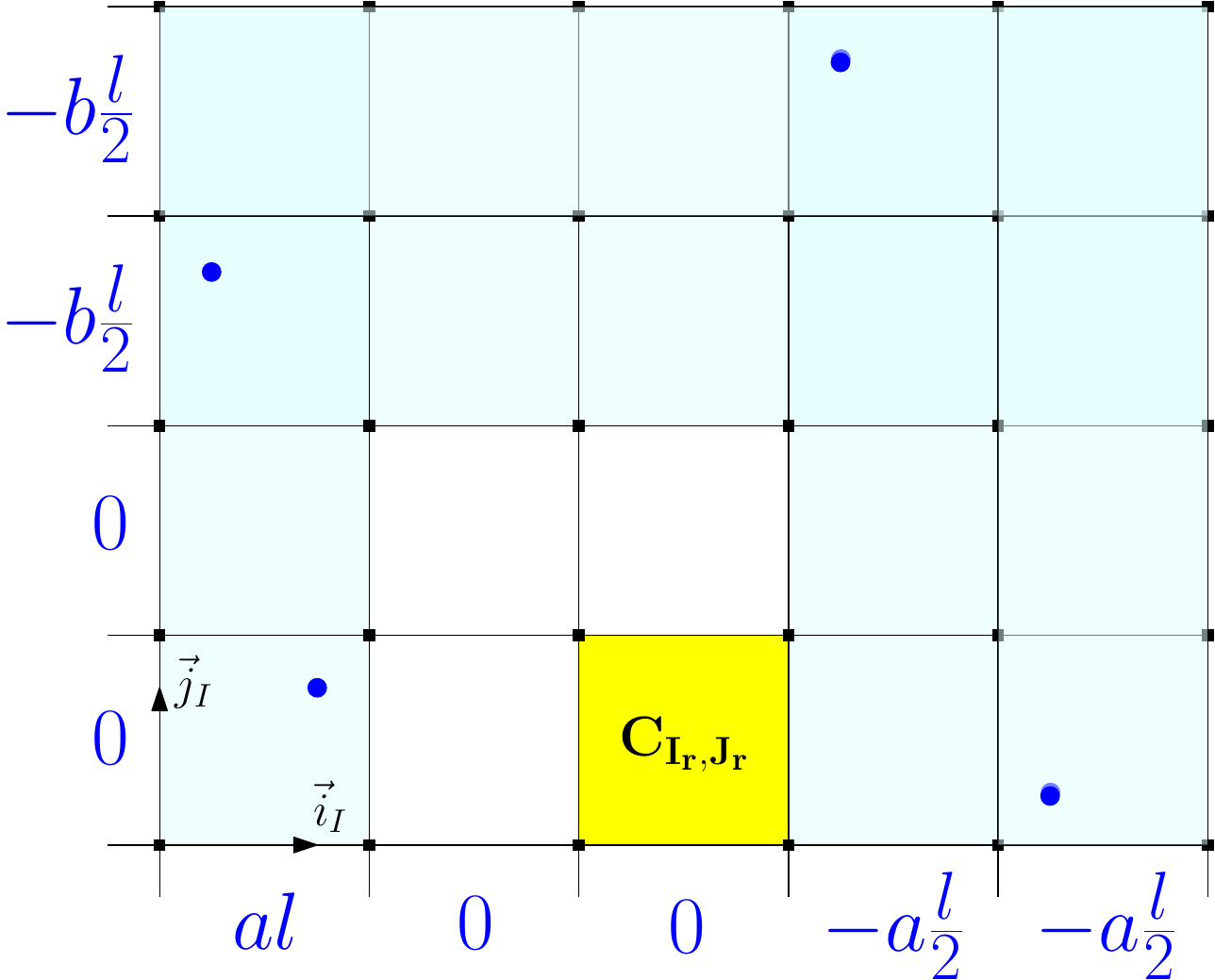}
\par\end{centering}
\caption{Example of the calculation of the sets $\bar{\mathcal{C}}$ and $\bar{\mathcal{R}}$
for an $\mathcal{S}\mathtt{(5,4)}$ surface. These sets are illustrated
by the lightly shaded rows and columns. The dots represent the location
of the objects on the surface. The reference cell is $C_{I_{r},J_{r}}= C_{3,1}$.
The horizontal and vertical numerical entries show the values of $\Delta Z_{1}^{I}$
and $\Delta Z_{2}^{J}$ for the distributed allocation control algorithm.\label{fig:example_distributed}}
\end{figure}

\begin{figure}
\centering
\includegraphics[scale=0.6]{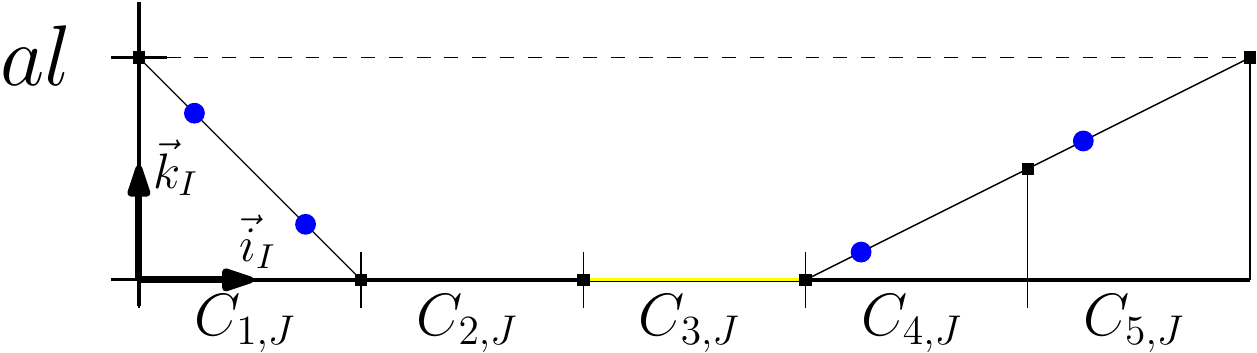}
\caption{Side view in the $\vec{i}_{I}$ direction of the $\mathcal{S}\mathtt{(5,4)}$
surface using the distributed allocation control law. \label{fig:View_I_Dist} }
\end{figure}

\begin{figure}
\centering
\includegraphics[scale=0.6]{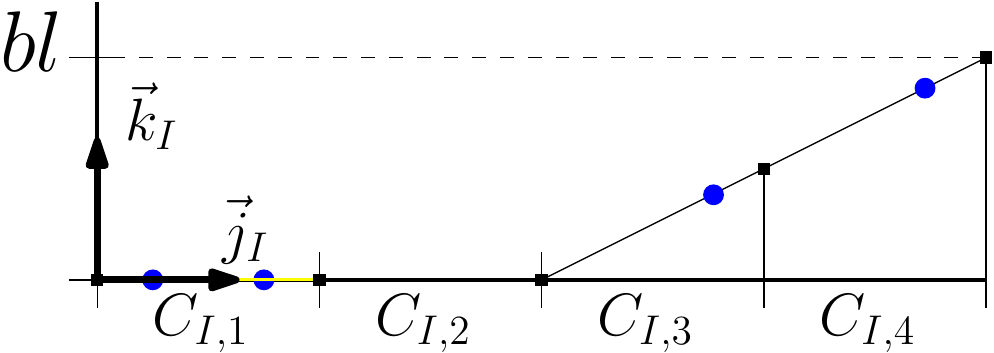}
\caption{Side view in the $\vec{j}_{I}$ direction of the $\mathcal{S}\mathtt{(5,4)}$
surface using the distributed allocation control law.\label{fig:View_J_Dist} }
\end{figure}

Since the total length of the actuators is fixed, the constants $a$ and $b$
represent the percentile of the total height that is allocated on the
$\vec{i}_{i}$ and $\vec{j}_{i}$ directions of the surface.
After the height differences are calculated, the individual actuator
lengths are updated starting with $Z_{i}^{I_{r},J_{r}}=0$ for $i=1,\ldots,4$.
It is important to note that the sets $\bar{\mathcal{C}}$ and $\bar{\mathcal{R}}$
change with time as objects are transported about the surface. An example
illustrating the calculation of $\Delta Z_{1}^{I}$ and $\Delta Z_{2}^{J}$
for an $\mathcal{S}\mathtt{(5,4)}$ surface at a given time instant
can be viewed in Figures \ref{fig:example_distributed}, \ref{fig:View_I_Dist} and \ref{fig:View_J_Dist}. Using this control scheme, it is possible to allot the inclination of the surface evenly to all objects, without wasting available height for empty cells.

\subsection{Wave Control Logic\label{sub:Wave-Control-Logic}}

The second control algorithm generates a ripple in the surface at the
rows and columns where the outermost objects are located. This ripple
converges like a wave towards the target location over time as the objects
gradually converge on the target cell. The rows and columns where
the ripple takes place have the maximum incline. This drives the furthest objects to the target and picks up the rest of them along the way. Using the same notation as described in the previous section, the height differences $\Delta Z_{1}^{I}$
($I\in\mathcal{C}$) and $\Delta Z_{2}^{J}$ ($J\in\mathcal{R}$)
are given by:
\begin{align*}
\Delta Z_{1}^{I}&=\begin{cases}
a\cdot l & \text{if }I=\min\left(\bar{\mathcal{C}_{l}}\right)\\
-a\cdot l & \text{if }I=\max\left(\bar{\mathcal{C}_{r}}\right)\\
0 & \text{else}
\end{cases} \\
\Delta Z_{2}^{J}&=\begin{cases}
b\cdot l & \text{if }I=\min\left(\bar{\mathcal{R}_{d}}\right)\\
-b\cdot l & \text{if }I=\min\left(\bar{\mathcal{R}_{r}}\right)\\
0 & \text{else}
\end{cases}
\end{align*}

\begin{figure}
\centering
\includegraphics[scale=0.3]{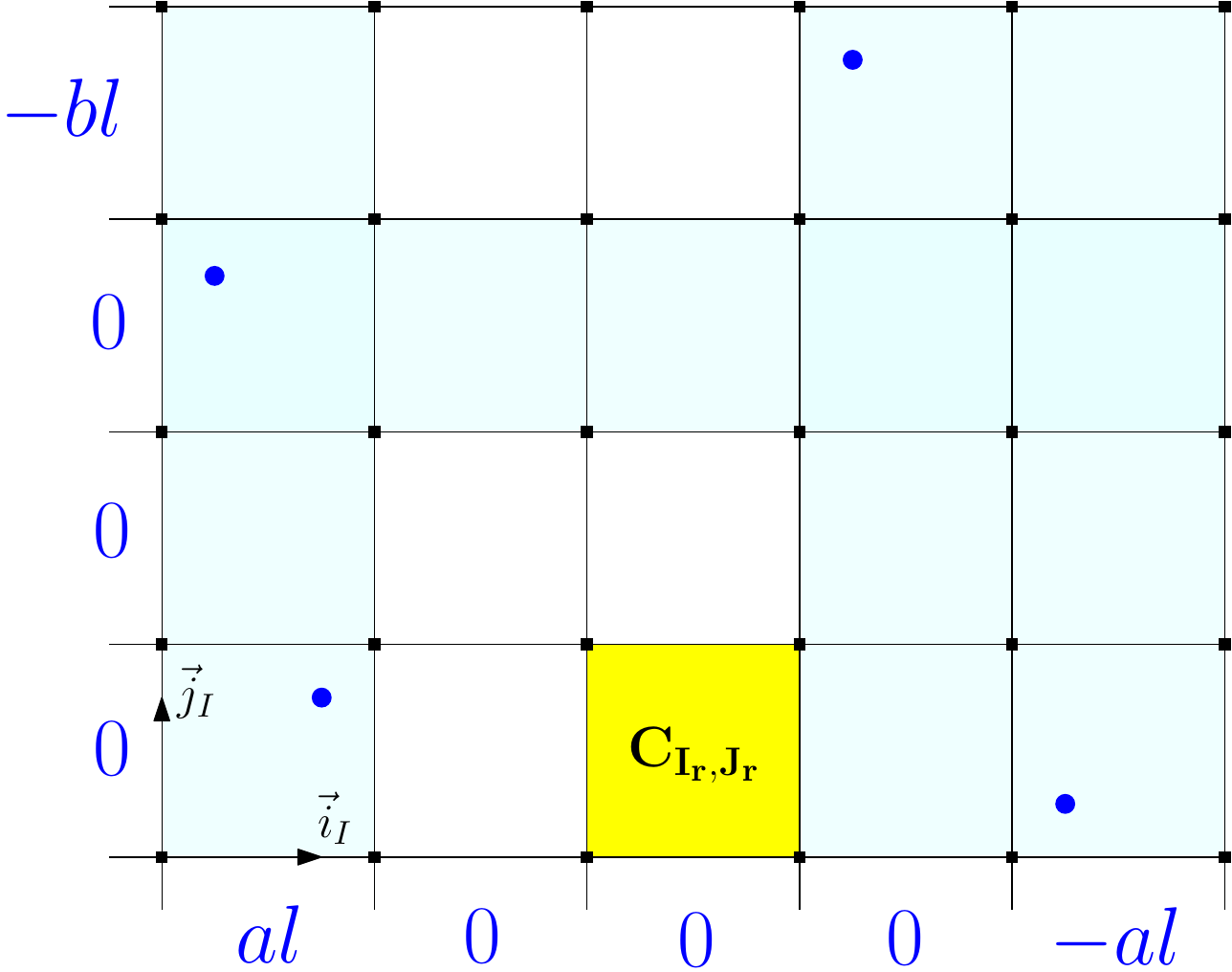}
\caption{The numerical entries show the values of $\Delta Z_{1}^{I}$ and $\Delta Z_{2}^{J}$
generated by the wave control algorithm for the example of the $\mathcal{S}\mathtt{(5,4)}$
surface.\label{fig:wave top view}}
\end{figure}

\begin{figure}
\centering
\includegraphics[scale=0.6]{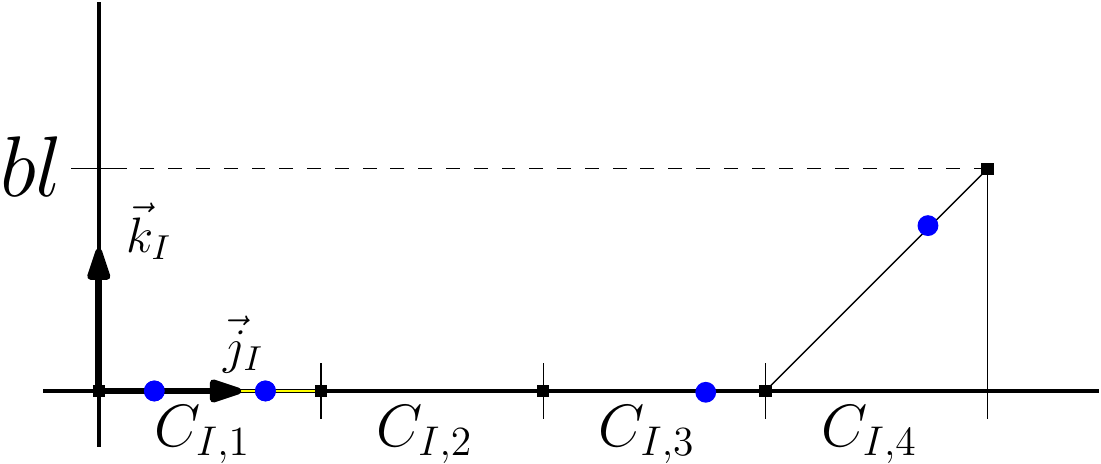}
\caption{Side view in the $\vec{j}_{I}$ direction of the $\mathcal{S}\mathtt{(5,4)}$
surface using the wave control law.\label{fig:View_J_wave} }
\end{figure}

The surface morphology observed during the execution of the wave control
algorithm is depicted in Figures \ref{fig:wave top view}, \ref{fig:View_I_wave} and \ref{fig:View_J_wave}. The wave
control algorithm applies the maximum possible force to the outermost
objects to gradually drive them collectively towards the target.

\subsubsection{Length Update Control Law}
After the height differences are calculated, the final step of the collective control algorithm involves the update of the individual actuator lengths. To ease the analysis the height of each actuator is defined as the sum $z^{I_{a},J_{a}}=z_{a}^{I_{a}}+z_{a}^{J_{a}}$ where $I_{a}=\left\{ 1,\ldots,n+1\right\} $ and $J_{a}=\left\{ 1,\ldots,m+1\right\}$ are the actuator columns and rows identifiers within the array. The pair $\left(z_{a}^{I_{a}},z_{a}^{J_{a}}\right)$ represent the length components of the actuator that are
working to drive the object in the $\vec{i}_{i}$ and $\vec{j}_{i}$ directions. The two components $z_{a}^{I_{a}}$ and $z_{a}^{J_{a}}$ have a constant value for every column $I_{a}$ and row $J_{a}$. The update law is initiated by leveling down the actuators of the reference cell $C_{I_{r},J_{r}}$. This action is accomplished by setting $z_{i}^{I_{i},J_{i}}=0$ for $I_{i}=\{I_{r},I_{r+1}\}$ and $J_{i}=\{J_{r},J_{r+1}\}$. The rest of the height components can be successively calculated using the following piecewise functions:
\begin{figure}
\centering
\includegraphics[scale=0.6]{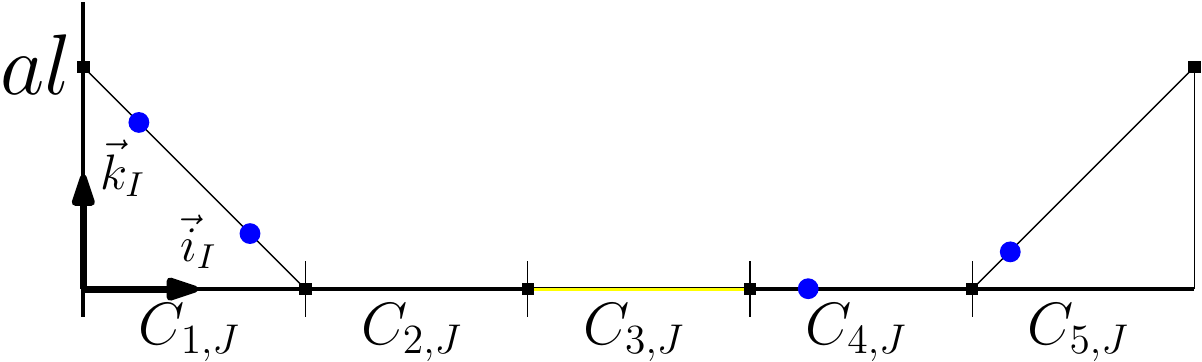}
\caption{Side view in the $\vec{i}_{I}$ direction of the $\mathcal{S}\mathtt{(5,4)}$
surface using the wave control law.\label{fig:View_I_wave} }
\end{figure}

\begin{equation}
z_{a}^{I_{a}}=\begin{cases}
\begin{array}{c}
-\sum_{I_{r}+1}^{I_{a}-1}\Delta Z_{1}^{k}\\
\sum_{I_{a}}^{I_{r}}\Delta Z_{1}^{k}
\end{array} & \begin{array}{c}
I_{a}\geq I_{r}+1\\
I_{a}\leq I_{r}
\end{array}\end{cases}\label{eq:Z_IA_A}
\end{equation}

\begin{equation}
z_{a}^{J_{a}}=\begin{cases}
\begin{array}{c}
-\sum_{J_{r}+1}^{J_{a}-1}\Delta Z_{2}^{k}\\
\sum_{J_{a}}^{J_{r}}\Delta Z_{2}^{k}
\end{array} & \begin{array}{c}
J_{a}\geq J_{r}+1\\
J_{a}\leq J_{r}
\end{array}\end{cases}\label{eq:Z_JA_A}
\end{equation}

\section{Experimental Prototype\label{sec:prototype}}

In this Section we outline in detail the electromechanical design
of a prototype conveying surface that autonomously morphs its shape
to transport a set of objects to a variable reference location. This
benchmark serves as an enabler to further our understanding related
to the kinematic capabilities and implementation limitations of LSANs.
The prototype system is an elastic surface with solid panels that
changes its shape by adjusting the heights of ten linear actuators.
The actuators placement forms four square cells in a `T' shaped configuration.
The `T' shape was chosen for the benchmark experimental system since
it requires the fewest actuators while still allowing the testing
of multiple different path elements.

The testbed was constructed on a pegboard base to position all of
the components. Each actuator was held in a vertical position using
two threaded rods to provide stability and support. The actuator model
used was a Frigelli L12-I with a $100\:mm$ stroke. The Frigelli L12
actuators were selected due to their low cost and relative high speed
($23\:mm/s$). A relative low time constant allows the actuators to
have a faster feedback response in order to control the moving objects
more effectively. Despite the relative low time constant of the selected
actuators, there is still a distinct time scaling between the actuators
and the objects dynamics. The Frigelli actuators gave a reduced strength
($43\:N$), however, this individual net force was still sufficient
for the network to collectively adjust the shape of the surface. 

\begin{figure}
\begin{centering}
\includegraphics[width=0.5\columnwidth]{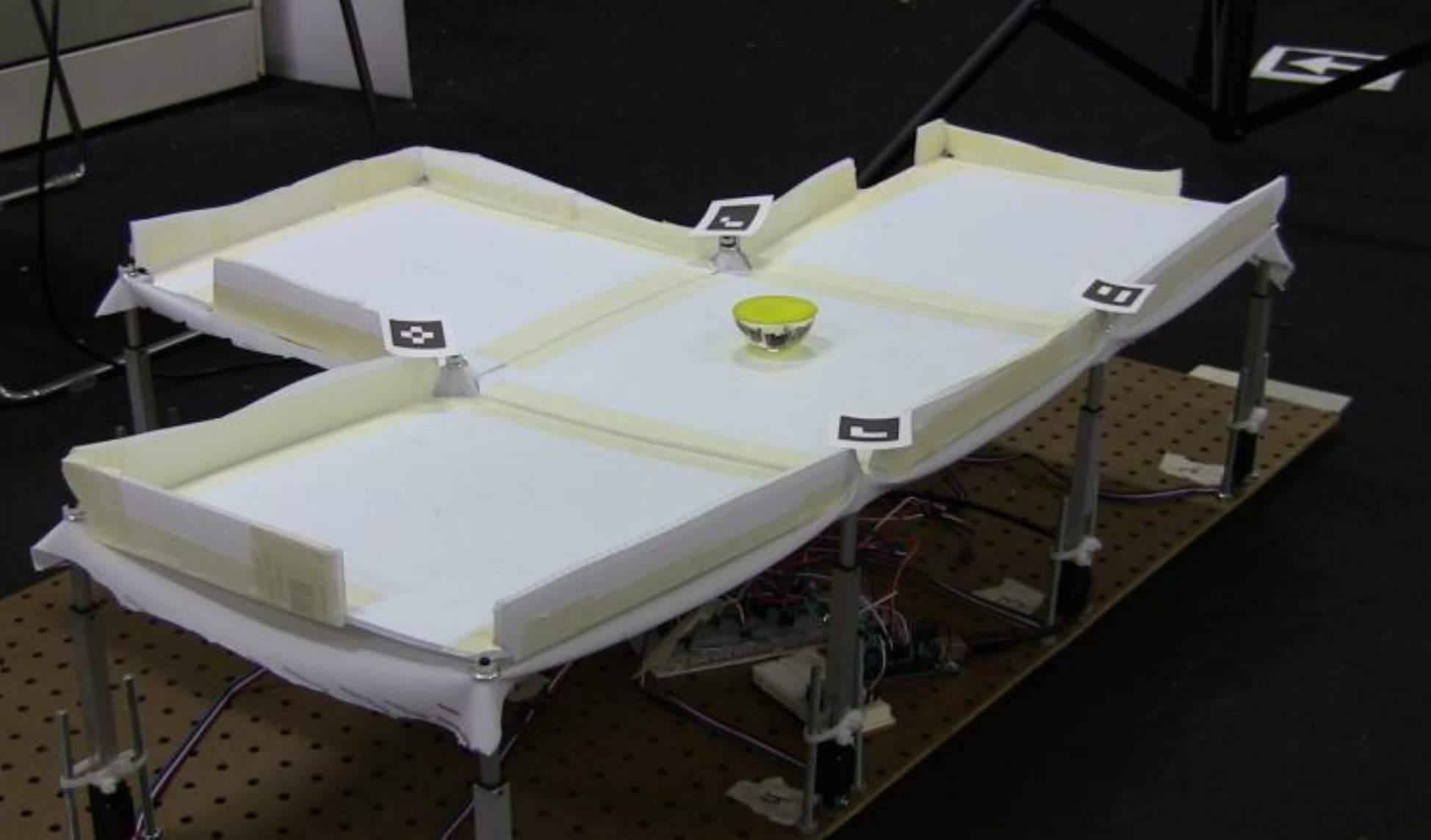}
\par\end{centering}
\caption{The experimental platform.\label{fig:1-1} }
\end{figure}

A sheet of spandex (10\%) and polyester (90\%) was mounted to the
tips of the actuators providing compliance for their length variation,
as well as securing the surface panels. Corrugated plastic surface
panels were placed on top of the spandex sheet due to their low coefficients
of static and kinetic friction. This property allows for object movement
at the expected slopes. The corrugated plastic was chosen after unsuccessfully
experimenting with multiple other materials ranging from neoprene
to latex. The object itself was selected after testing many different
candidates ranging from wood cubes to Ping-Pong balls. The object
(seen in Figure \ref{fig:1-1}) that provided the most controllable
and repeatable motion was a polycarbonate half sphere. 

The system was controlled using an Arduino Uno microcontroller. This
microcontroller was selected for its sufficient processing power (able
to execute the control algorithm at $11Hz$), its low cost, and most
importantly its minimal development time. The Arduino Integrated Development
Environment (IDE) allows the rapid development and testing due to
the large number of available low level execution libraries of various
components (e.g. servos and motors). The system's operation with a
simple and inexpensive controller is an ideal way to demonstrate the
applicability of the proposed mechanism. 

The longitudinal/lateral position measurements of the objects were generated
initially by an external and commercially available visual tracking
system called Roborealm and later on by a custom program developed
in OpenCV. The vision sensor itself was a simple Live! Cam inPerson
HD VF0720 web-camera that was placed on top of the surface. The video
feed from the overhead web-cam is processed by the tracking system
that calculates the planar coordinates of the object. These coordinates
are the feedback signal required by the control algorithm. Data was
sent from the computer to the Arduino micro-controller via a serial
connection and an XBee wireless RF module at $7Hz$. The assembled
surface apparatus with the vision sensor can be seen in Figure \ref{fig:1-1}. 

\section{Results\label{sec:results}}
\subsection{Numerical Simulations\label{sub:Multi-Object-control}}
This Section provides an evaluation of the proposed multi-cell control algorithms via extensive numerical simulations. The two algorithms (distributed allocation and wave) were compared with a benchmark static funnel configuration using a Matlab/Simulink model. The simulation used 20 objects starting from random locations on a $\mathcal{S}(5,6)$ rectangular surface. The reference cell was chosen to be $C_{I_{r},J_{r}}=C_{3,1}$. For the simulation runs, the same set of parameters (length, width, and initial positions) was used for all three cases so the results could be directly compared. The length and width of each cell is set to $L=200\,cm$ and $W=200\,cm$, respectively. The actuators' stroke was set to $l=100\,cm$. Each object was modeled with a mass of $m=1\,kg$, and the friction coefficient between the object and the surface is set to $b_{f}=0.1$. The simulation was executed by numerically integrating Newton's equation of motion for each object (Section \ref{sub:Object-Dynamics}). A barrier is assumed at the exterior margins of the surface to keep the objects confined in the workspace of the mechanism. The impacts of the objects with the barrier are modeled as elastic such that no energy is dissipated from the collision. 

\begin{figure}
\centering
\includegraphics[width=0.7\columnwidth]{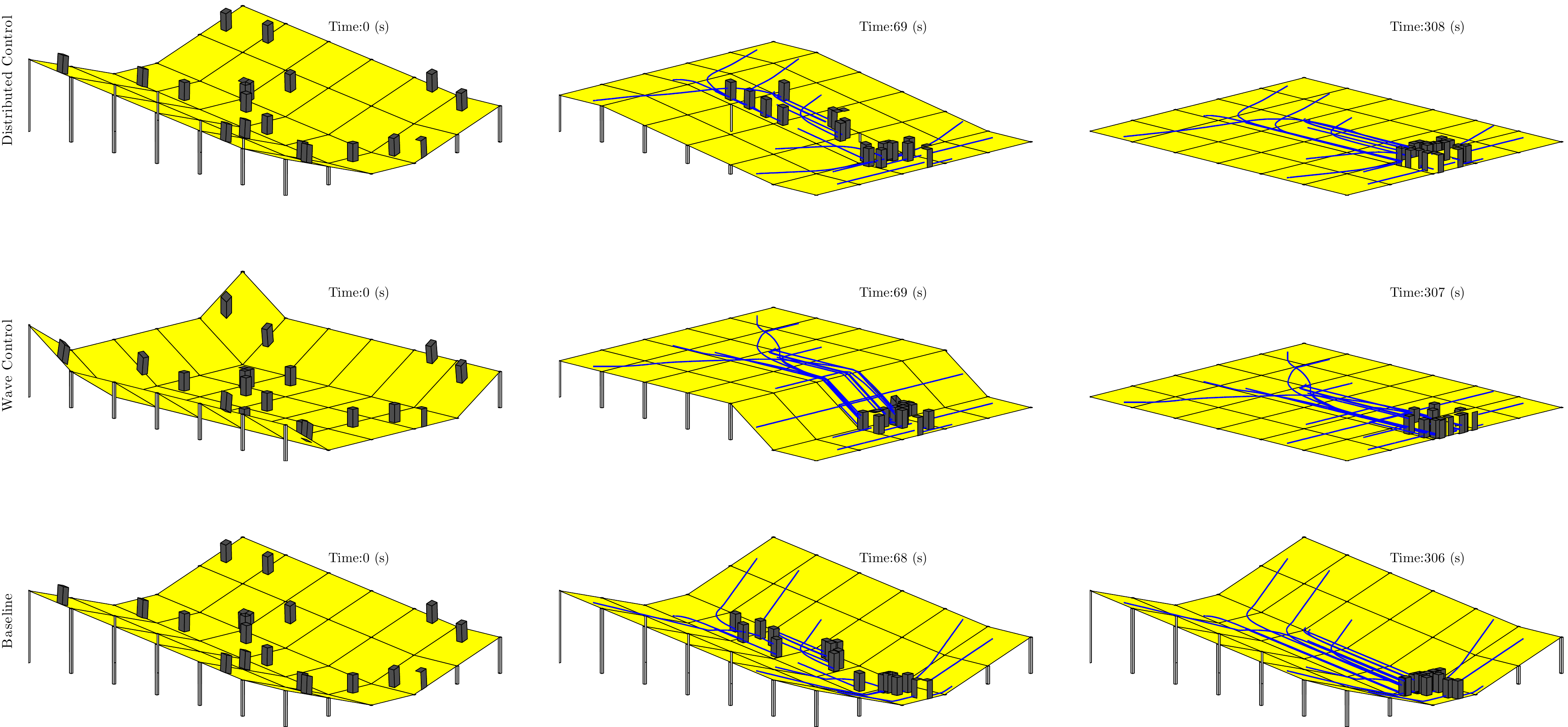}
\caption{Animation of the surface for the three control algorithms at three time instances.\label{fig:animation_V1}}
\end{figure}

The configuration of the surface for the three distinct algorithms (distributed allocation, wave and static funnel) are animated for different time instances in Figure \ref{fig:animation_V1}. It can be seen that at least one object was placed initially in each row and column of the surface. This causes the distributed allocation control algorithm (for the initial time instant) to spread the available incline across all of the cells, resulting in a even distribution of the slope, identically to the static funnel case. 

\begin{figure}[t]
\centering
\includegraphics[width=0.7\columnwidth]{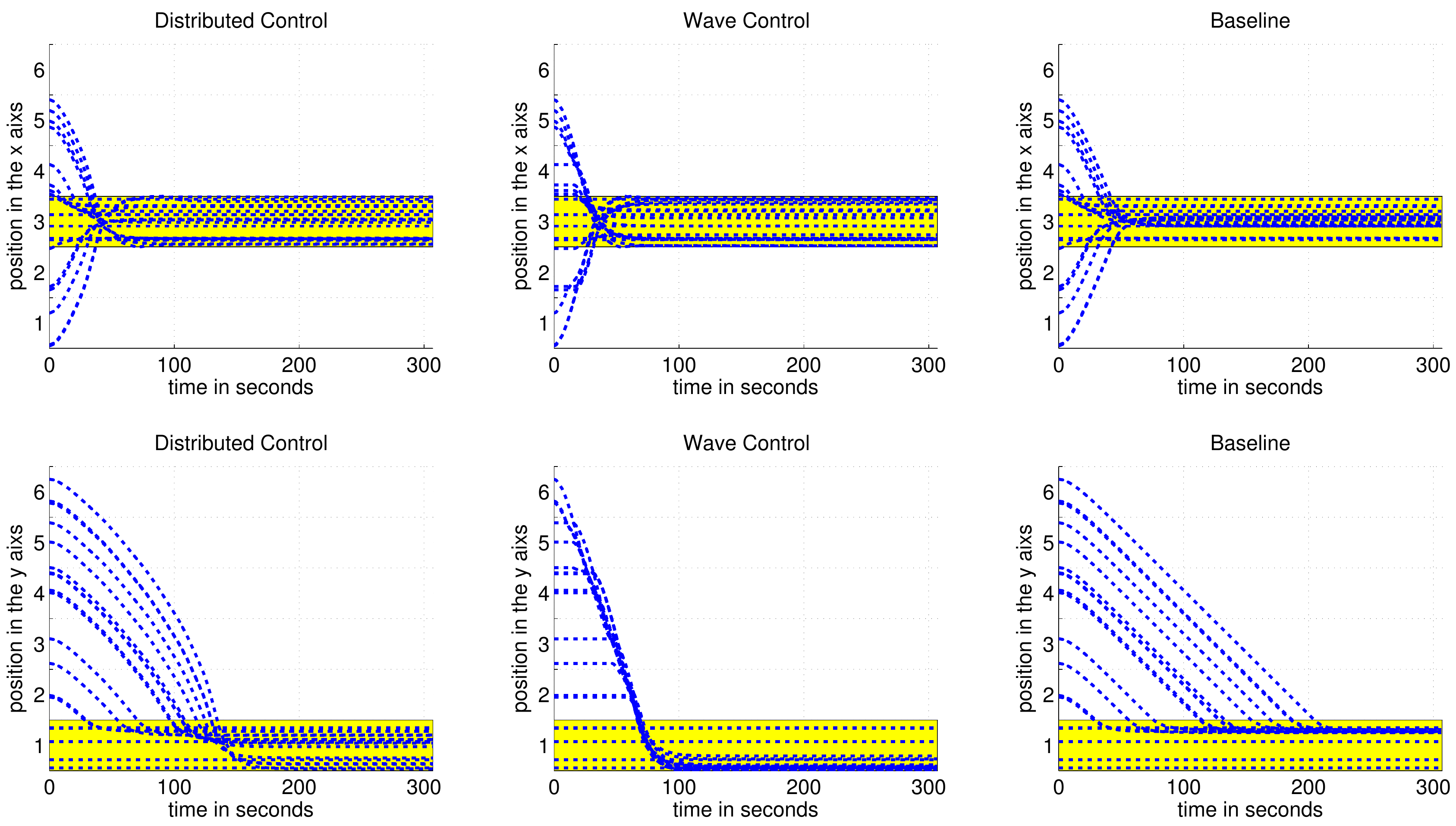}
\caption{Objects position with respect to time. The shadowed area represents the length of the reference cell. \label{fig:Plots-of-object_V1}}
\end{figure}

The locations of the objects in the $\vec{i}_{I}$ and $\vec{j}_{I}$ directions, for the three algorithms, with respect to time are illustrated in Figure \ref{fig:Plots-of-object_V1}. By inspection, it can be seen that in both directions the wave control algorithm delivers the objects to the reference location in the shortest total time. The baseline funnel exhibits the slowest overall response. The wave algorithm takes approximately $70\,sec$, the distributed algorithm $130\,sec$, and the baseline funnel takes $200\,sec$ to converge. The wave control algorithm injects to the objects maximal amounts of potential energy through the largest inclinations of the outermost cells, resulting to the highest kinetic energy conversion. The wave control has the additional benefit that all of the objects arrived to the final destination at approximately the same time.

\begin{figure}
\centering
\includegraphics[width=0.7\columnwidth]{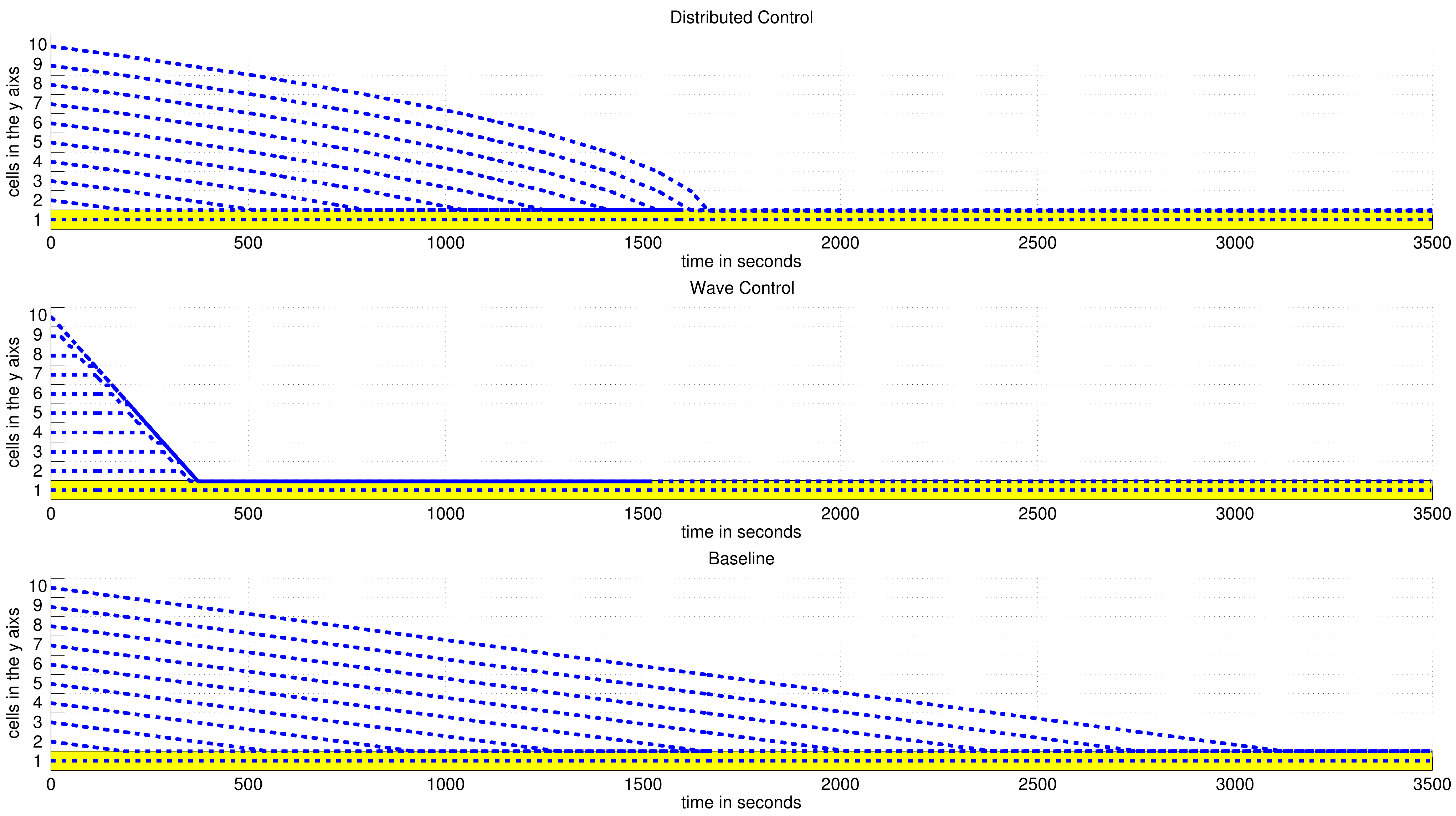}
\caption{Comparison of position with respect to time for the three control
algorithms in the $\vec{i}_I$ direction of the $\mathcal{S}(1,10)$ surface. \label{fig:Comparasom-of-position}}
\end{figure}

\begin{figure}
\centering
\includegraphics[width=0.7\columnwidth]{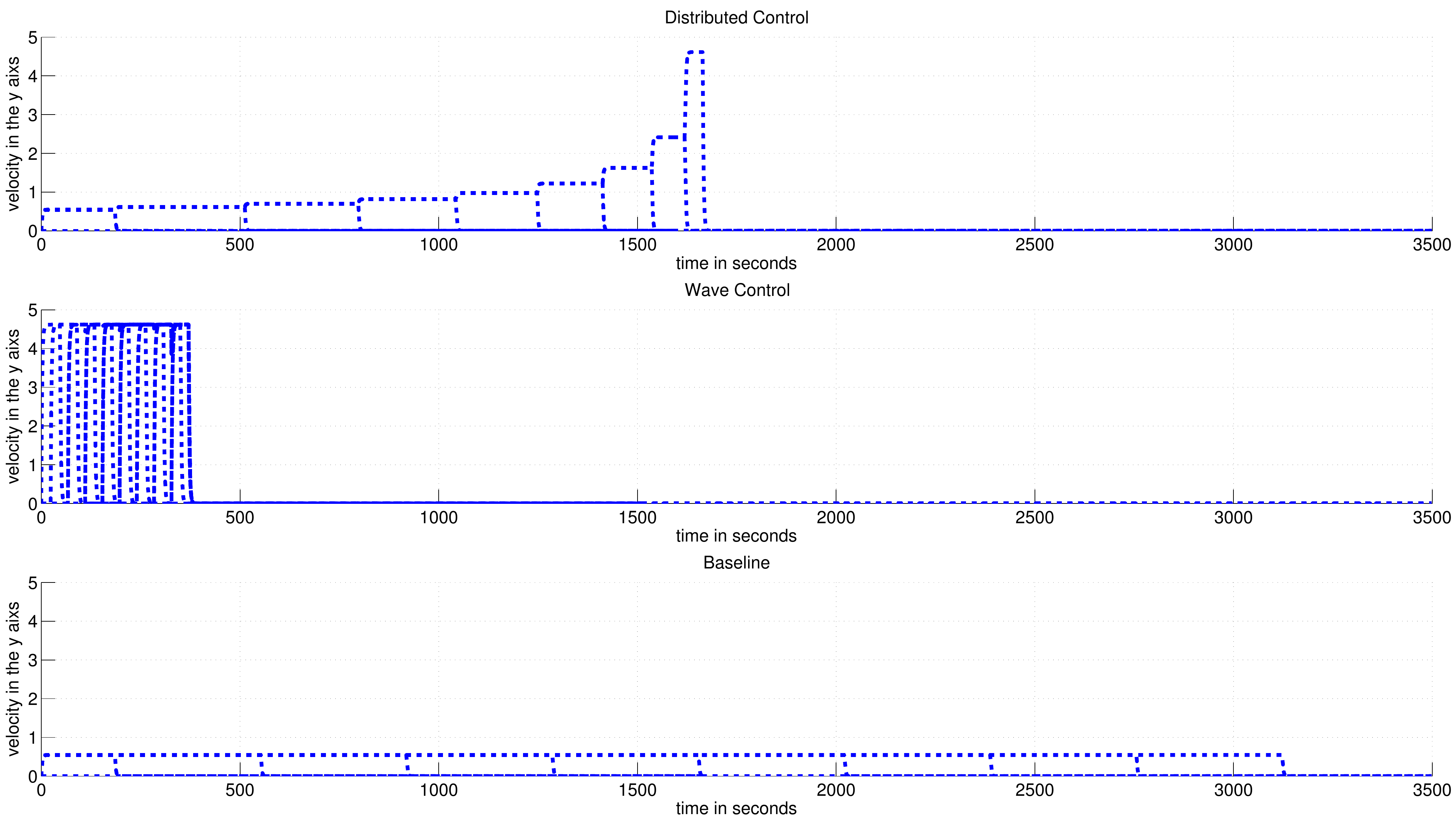}
\caption{Comparison of velocity with respect to time for the three control
algorithms in the $\vec{i}_I$ direction of the $\mathcal{S}(1,10)$ surface. \label{fig:Comparison_vel}}
\end{figure}

The behavior of each control algorithm can be better understood by investigating the special case of the single track surface $\mathcal{S}(1,10)$. In this scenario the reference cell is $C_{I_{r},J_{r}}=C_{1,10}$. The simulations are initiated having a single object lying to each cell of the surface. The motion of the objects is restricted to the $\vec{i}_I$ direction. All surface and objects parameters are the same with the first case study. The position and velocity of the objects in the $\vec{i}_{I}$ direction, for the three algorithms, with respect to time are depicted in Figures \ref{fig:Comparasom-of-position} and \ref{fig:Comparison_vel}, respectively. 

Obviously, the static inclination funnel (baseline) exhibits the slowest convergence time. In all cases the velocity of each object is determined by the slope of the cell that is  subsequently filtered by the low pass transfer function $1/\left(s+b_{f}\right)$. A large value for $b_{f}$ implies small final velocity and fast response. The steady state velocity of the objects is calculated by the term $(g/b_{f})C_{\theta}C_{\phi}S_{\theta}$ based on \eqref{eq:eom}. Therefore, maximum inclination results to maximum velocity. This is the velocity that the objects obtain gradually, during the execution of the wave algorithm, starting from the most remote ones. In the distributed allocation algorithm, the kinetic energy of the system increases only when cells are free of objects, resulting to a slower final convergence.    


\begin{figure}[t]
\centering
\includegraphics[width=0.35\columnwidth]{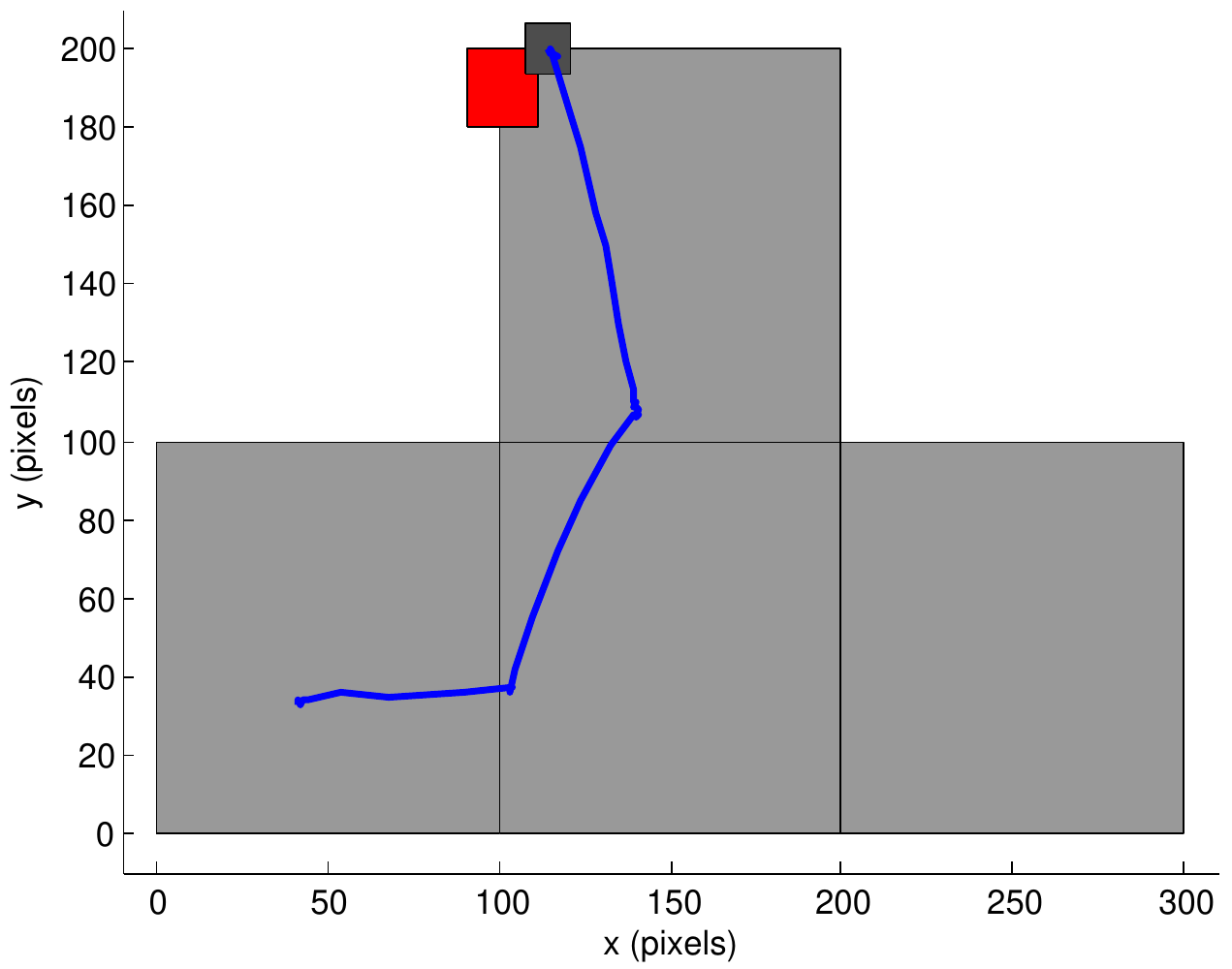}\includegraphics[width=0.35\columnwidth]{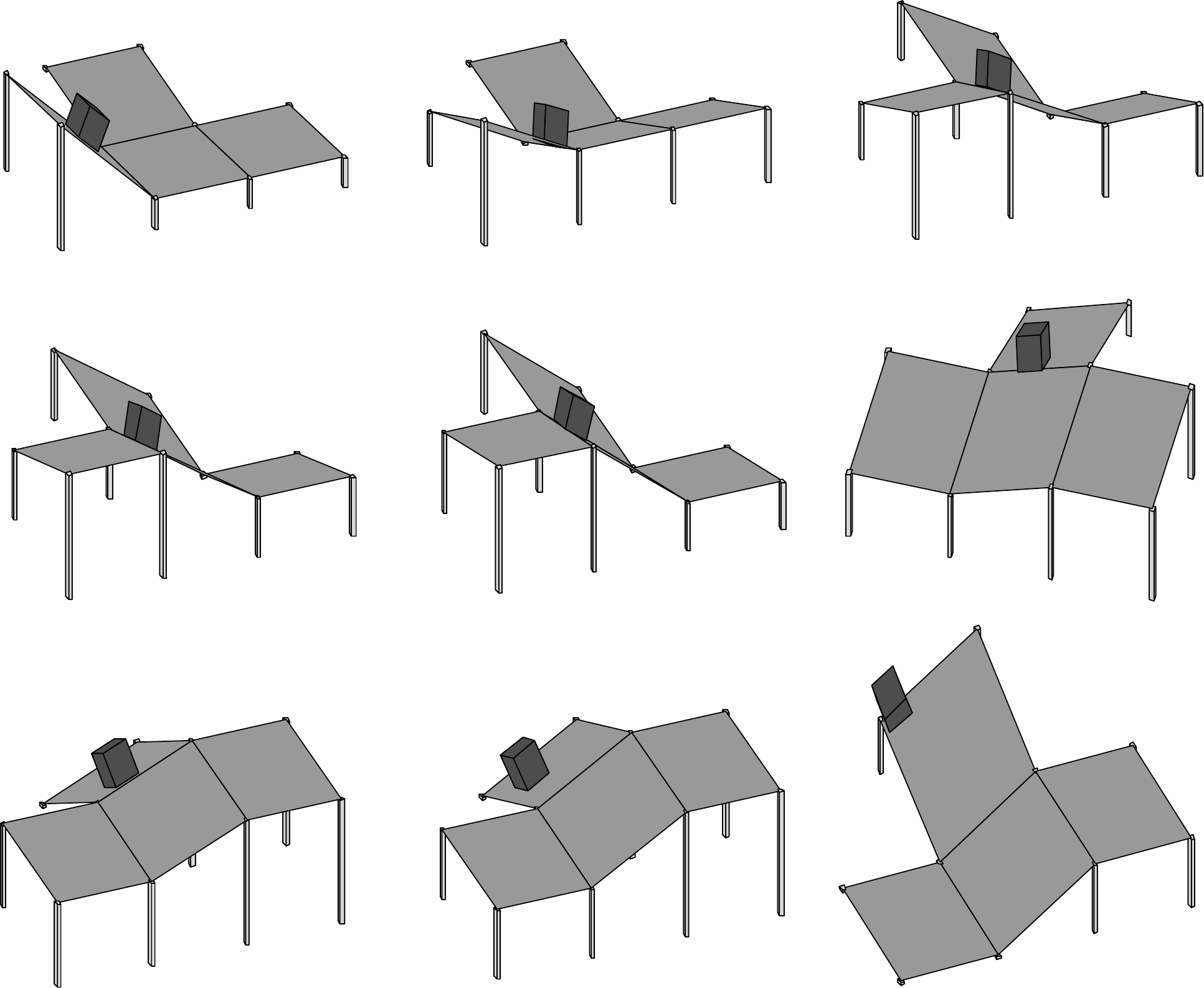}
\caption{2-D top view of the object's planar motion based on experimental data.
The panels of the surface are shown in light gray, the target zone
in red, the object is shown in dark gray, and the blue line is the
path that the object followed. The right side illustrates the 3-D
visualization of the surface and the object for nine different time
instances based on experimental data.\label{fig:8}}
\end{figure}

\subsection{Experimental Results\label{sub:Prototype-results}}
The wave control algorithm was further validated using the prototype system described in Section \ref{sec:prototype}. The multi-cell surface was able to move the object consistently and successfully to an exit point on its designated reference cell. The tested path was primarily chosen to demonstrate the basic movements that are possible on the surface. The reference path requires that the object makes a U-turn across the cells to reach the target exit point. This case represents a real-world scenario where the object has to be transported around a hole or a wall. A two dimensional top view of the object's actual trajectory based on experimental data is shown in the left side of Figure \ref{fig:8}. The right side of the same Figure illustrates the surface's morphology at different time instances of the test run. The object successfully reached the reference location by autonomously navigating across multiple cells. At every transition between adjacent cells, the object is moving over the midpoint of their common side.

The preliminary experiments that took place on the prototype mechanism demonstrated that the maximum slope (maximum propulsion force) is required to overcome the static friction and allow the object to slide when starting from rest. To this extent, the tuning parameters of the wave control algorithm $a$, $b$ (Section \ref{sub:Wave-Control-Logic}) are set to either zero or one providing full inclination in a single direction. This modification was deemed necessary to obtain greater slopes, thus, greater applied forces for moving the objects.
 
\begin{figure}[t]
\includegraphics[width=1.60in]{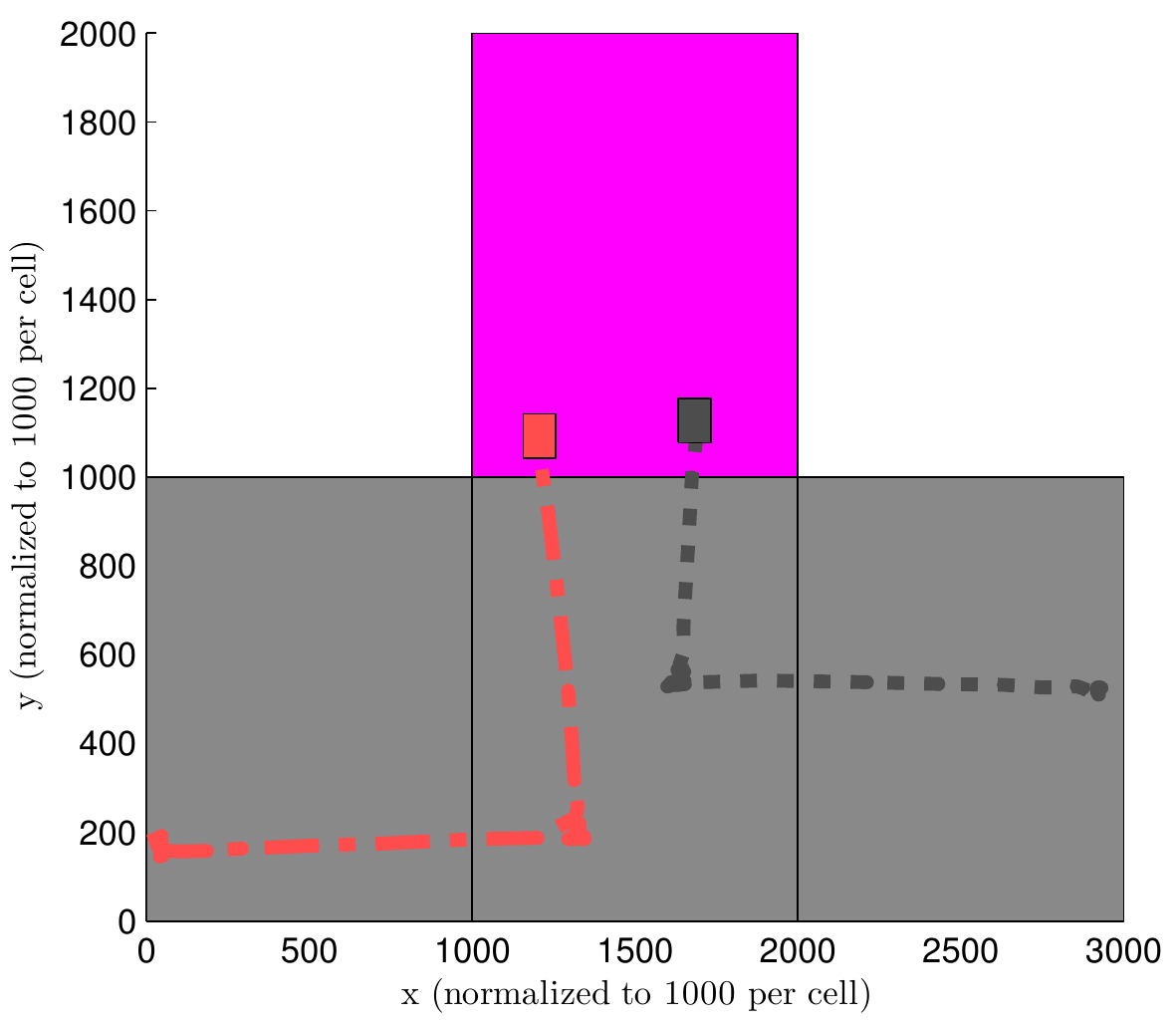}\includegraphics[width=1.60in]{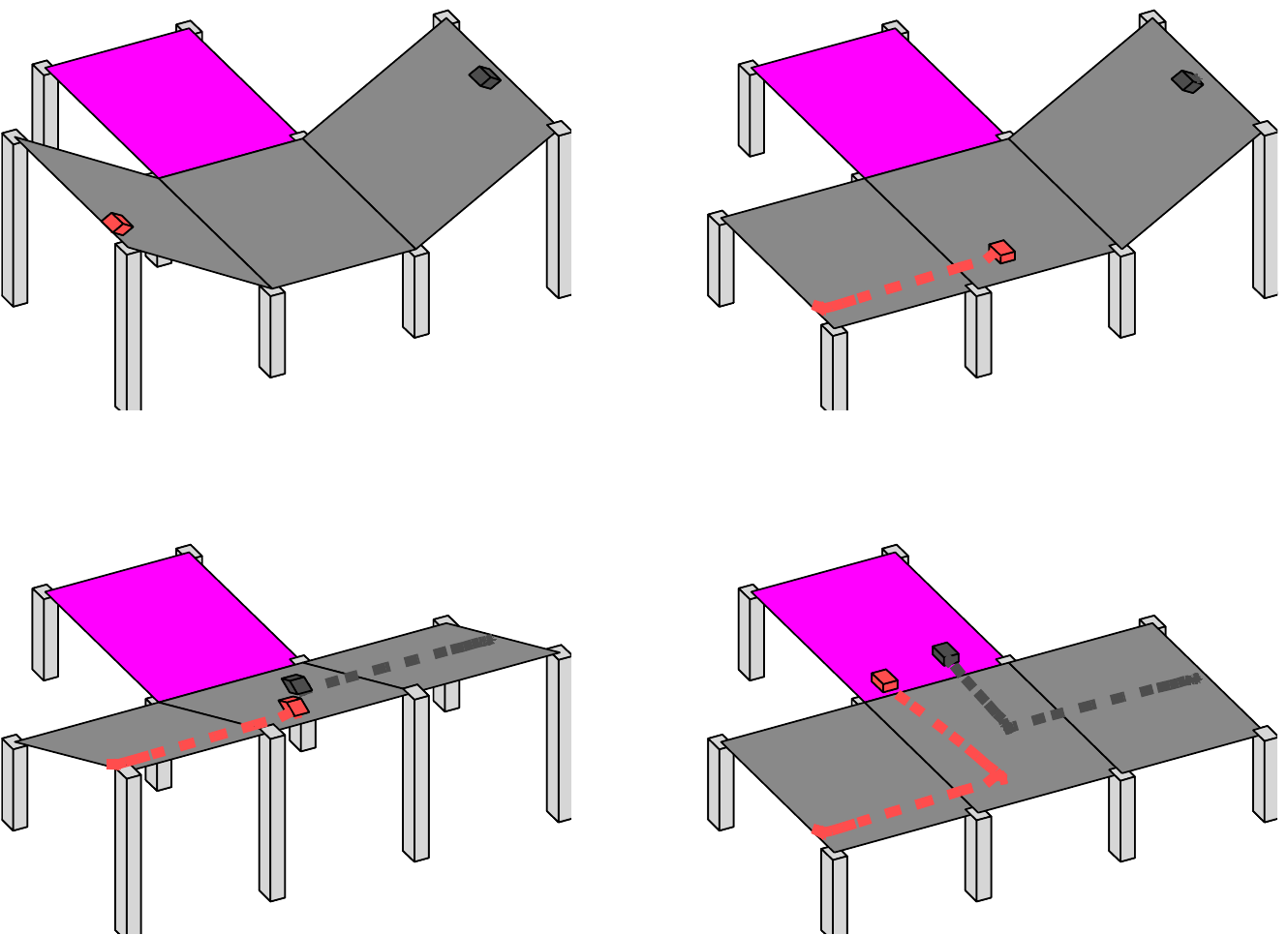}
\includegraphics[width=1.60in]{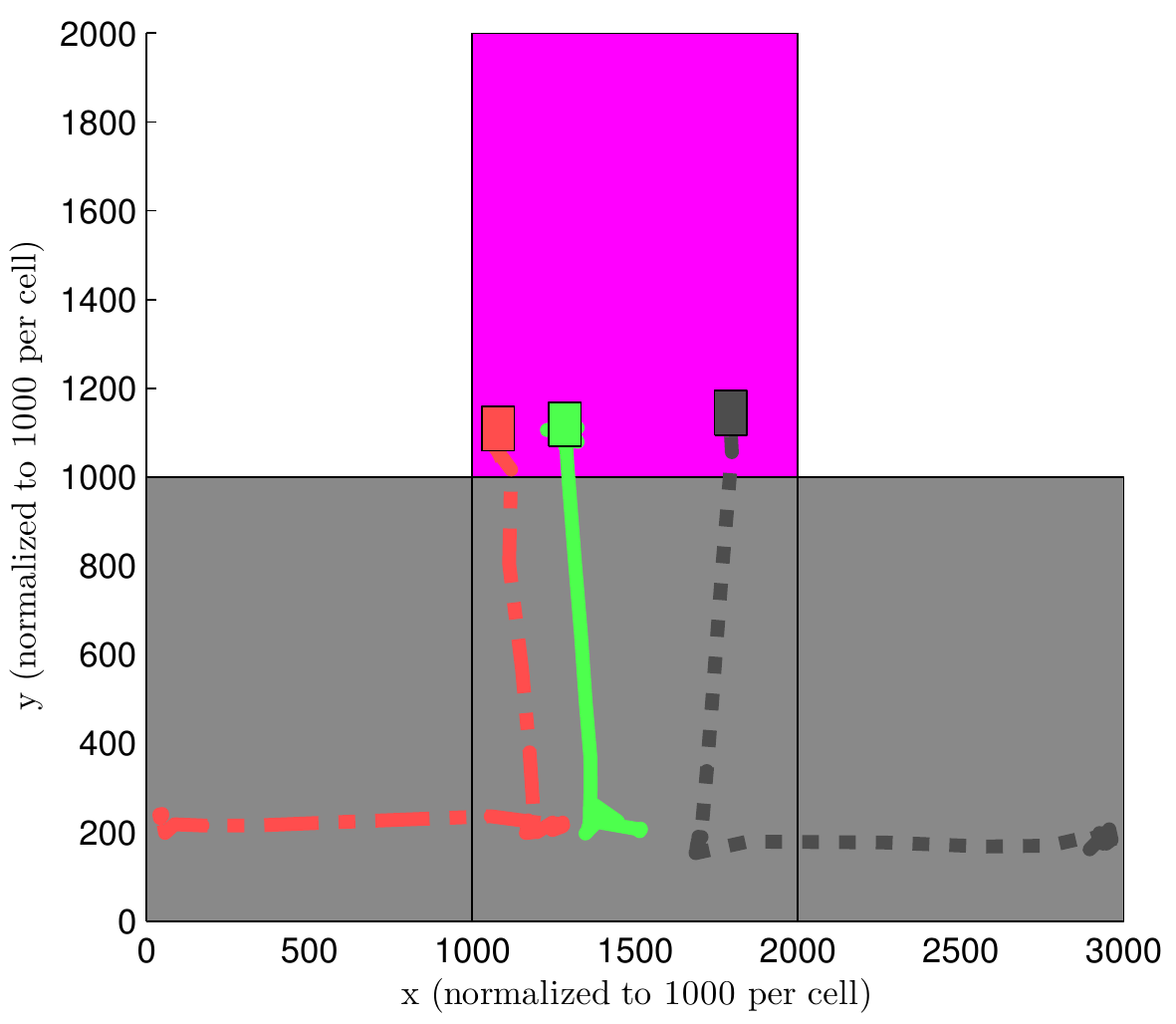}\includegraphics[width=1.60in]{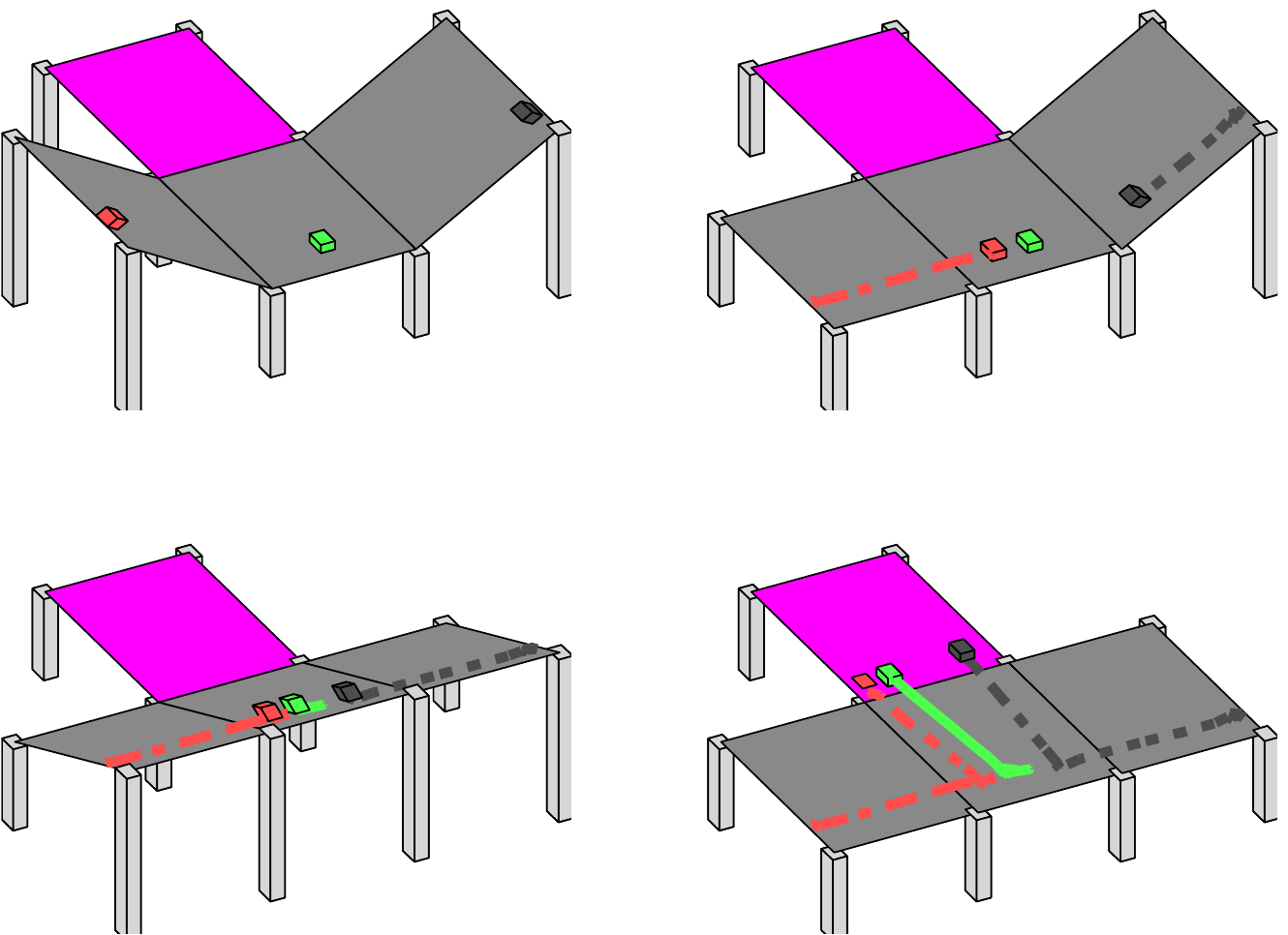}
\centering
\includegraphics[width=1.60in]{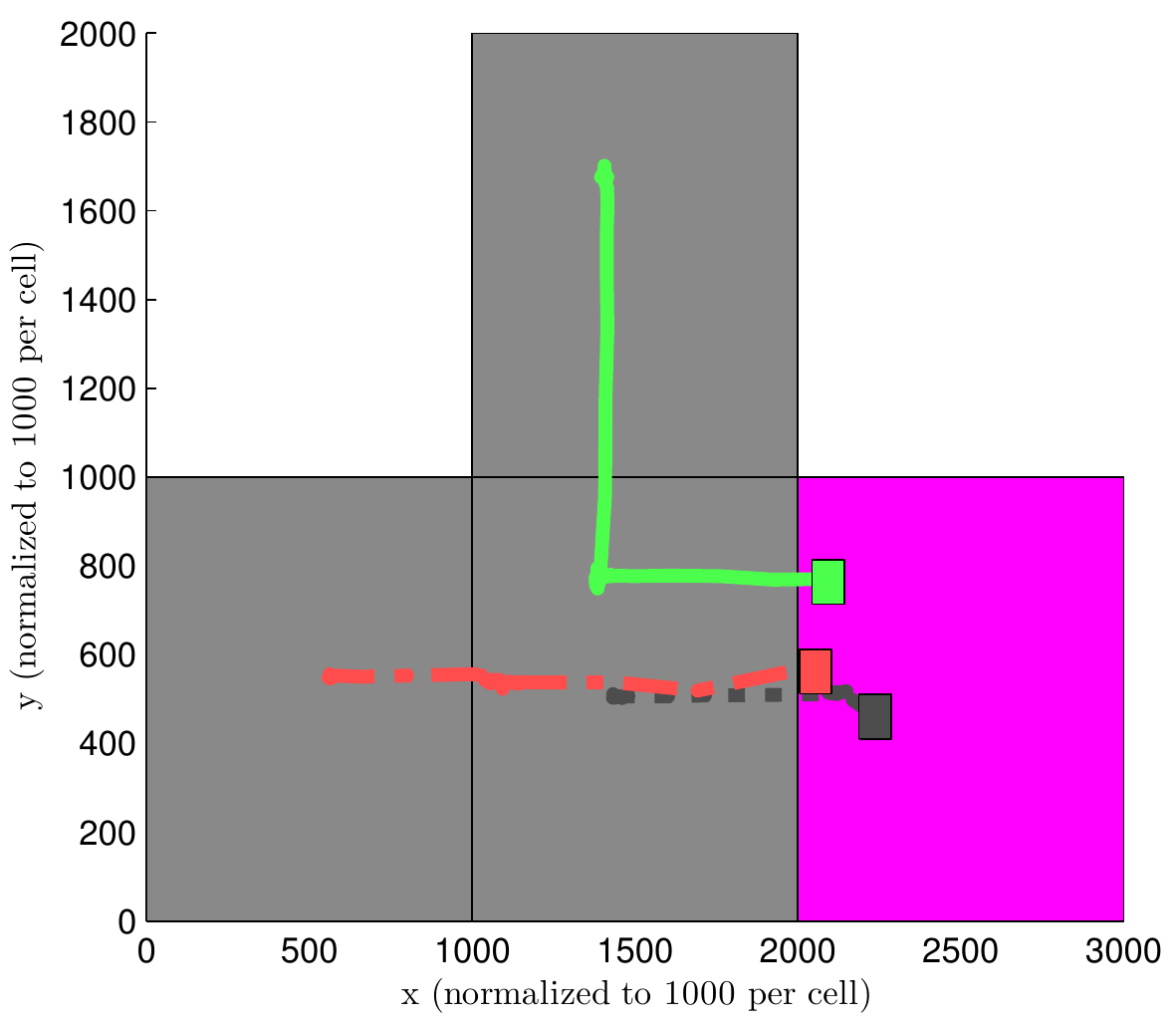}\includegraphics[width=1.60in]{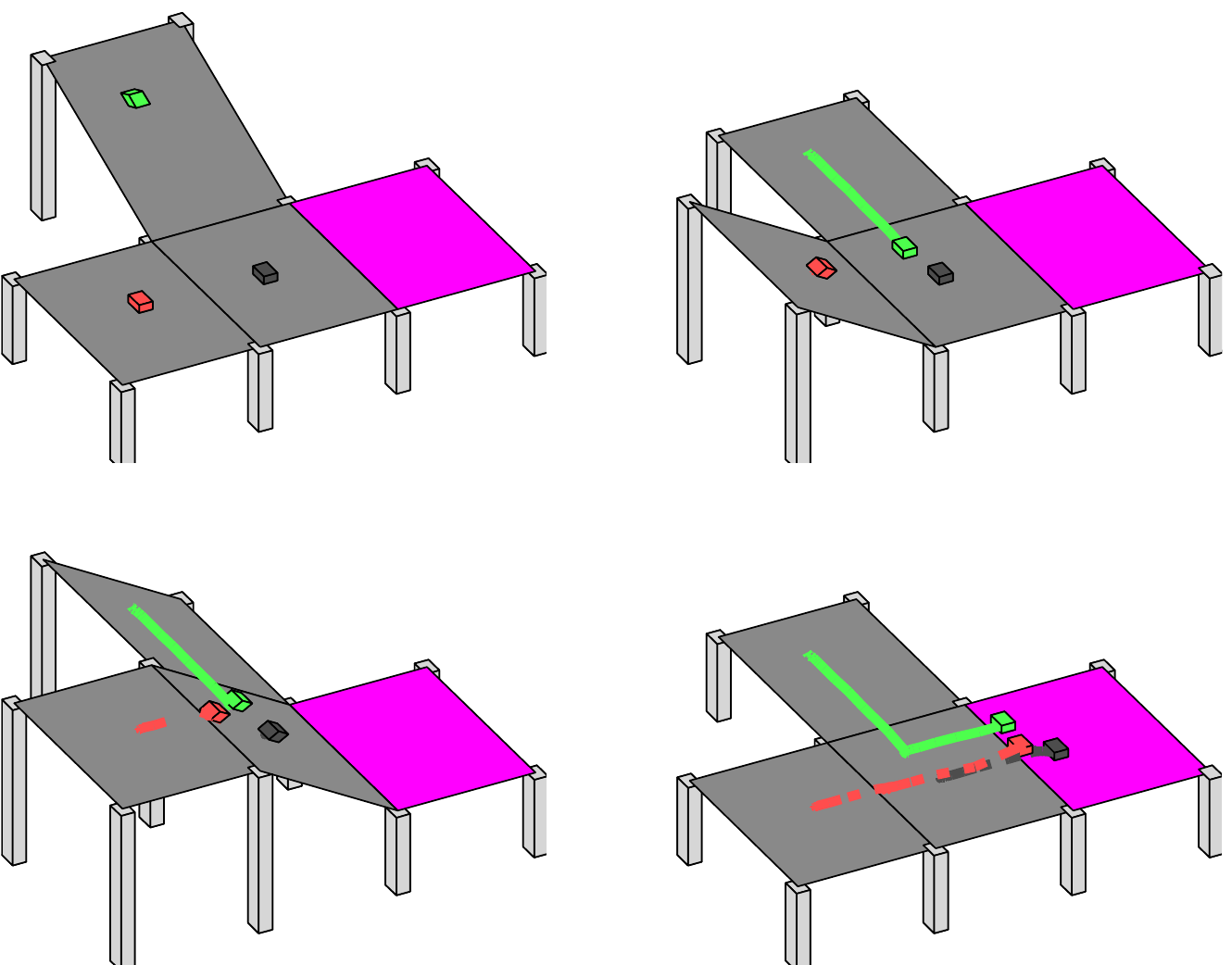}
\caption{2-D top view of the object's planar motion based on experimental data.
The panels of the surface are shown in light gray, the target zone
in magenta, the objects are color coded, and the lines are the path
that the object followed. The right side illustrates the 3-D visualization
of the surface and the object for four different time instances based
on experimental data.\label{fig:3object test L2}}
\end{figure}

The wave control algorithm was further tested using 2 objects with repeatable success. Sample test run are shown in Figure \ref{fig:3object test L2}. In this experiment two objects, starting from the outermost cells of the surface, are converging to to the center front reference cell. The final level of complexity involved three objects in multiple configurations. In all tested configurations the surface was able to repeatedly deliver the objects successfully to the reference location.

\section{Conclusion \label{sec:Conclusion}}
This work provides an analytic and experimental study of a LSAN for distributed manipulation. The presented mechanism involves a morphing surface that autonomously adjusts its shape to transport an arbitrary number of objects to a reference location. The morphing process takes place by a grid of linear actuators that adjust their height. A detailed analytical study of this mechanism is provided that results to an explicit calculation of the available control resources. The main focus of this work is the derivation of computationally attractive control algorithms that can handle efficiently an arbitrary number of actuators. A prototype testbed was developed by off-the-shelf components to validate the applicability of this originally concept and to reveal potential limitations that do not emerge from the theoretical analysis. Both control algorithms that were investigated in this work show significant improvements in performance when compared against the conventional static inclination solution.

\end{document}